\documentclass[lettersize,journal]{IEEEtran}
\usepackage{amsmath,amsfonts}
\usepackage{algorithmic}
\usepackage{algorithm}
\usepackage{array}
\usepackage[caption=false,font=normalsize,labelfont=sf,textfont=sf]{subfig}
\usepackage{textcomp}
\usepackage{stfloats}
\usepackage{url}
\usepackage{verbatim}
\usepackage{graphicx}
\usepackage{cite}
\usepackage{booktabs}  %% 三线表
\usepackage{diagbox}   %% 斜线表头
\usepackage{multirow}  %% 合并单元格

\begin{document}

\title{Global-Aware Enhanced Spatial-Temporal Graph Recurrent Networks: A New Framework For Traffic Flow Prediction}

\author{Haiyang Liu, Chunjiang Zhu, and Detian Zhang
        % <-this % stops a space
\thanks{Haiyang Liu and Detian Zhang are with the Institute of Artificial Intelligence, Department of Computer Science and Technology, Soochow University, Suzhou, China. (e-mail: 20215227052@stu.suda.edu.cn; detian@suda.edu.cn). Chunjiang Zhu is with the Department of Computer Science, University of North Carolina at Greensboro, Greensboro, NC, USA. (e-mail: chunjiang.zhu@uncg.edu).} 
%Jin Wang is with the Department of Computer Science and Technology, Soochow University, Suzhou, China. (e-mail: wjin1985@suda.edu.cn).}% <-this % stops a space
% \thanks{Chunjiang Zhu was supported by UNCG Start-up Funds and Faculty First Award. Detian Zhang is partially supported by the Collaborative Innovation Center of Novel Software Technology and Industrialization, the Priority Academic Program Development of Jiangsu Higher Education Institutions.}
\thanks{Corresponding author: Detian Zhang.}}

% The paper headers
\markboth{Journal of \LaTeX\ Class Files,~Vol.~14, No.~8, August~2021}%
{Shell \MakeLowercase{\textit{et al.}}: A Sample Article Using IEEEtran.cls for IEEE Journals}

\IEEEpubid{0000--0000/00\$00.00~\copyright~2021 IEEE}
% Remember, if you use this you must call \IEEEpubidadjcol in the second
% column for its text to clear the IEEEpubid mark.

\maketitle

\begin{abstract}
Traffic flow prediction plays a crucial role in alleviating traffic congestion and enhancing transport efficiency. While combining graph convolution networks with recurrent neural networks for spatial-temporal modeling is a common strategy in this realm, the restricted structure of recurrent neural networks limits their ability to capture global information. For spatial modeling, many prior studies learn a graph structure that is assumed to be fixed and uniform at all time steps, which may not be true. This paper introduces a novel traffic prediction framework, Global-Aware Enhanced Spatial-Temporal Graph Recurrent Network (GA-STGRN), comprising two core components: a spatial-temporal graph recurrent neural network and a global awareness layer. Within this framework, three innovative prediction models are formulated. A sequence-aware graph neural network is proposed and integrated into the Gated Recurrent Unit (GRU) to learn non-fixed graphs at different time steps and capture local temporal relationships. To enhance the model's global perception, three distinct global spatial-temporal transformer-like architectures ($\mathrm{GST^2}$) are devised for the global awareness layer. We conduct extensive experiments on four real traffic datasets and the results demonstrate the superiority of our framework and the three concrete models.

\end{abstract}


\section{Introduction}
\noindent With the increasing adoption of sensors like speed cameras and loop detectors, traffic management has become more efficient and convenient. These sensors collect vast amounts of traffic data (e.g., traffic flow and traffic speed) from various detection points.
Traffic flow prediction is crucial within Intelligent Transport Systems (ITS), utilizing historical traffic data for accurate traffic flow prediction. Accurate predictions help prevent congestion and minimize traffic-related costs. \cite{wu2020comprehensive,jiang2021dl}.

Traffic data possesses strong dynamic characteristics in both spatial and temporal dimensions. Effectively capturing the spatial-temporal dependencies and accurately predicting future information from non-linear traffic data is a major challenge in traffic prediction. Currently, spatial-temporal graph neural networks (STGNNs) are widely used in the field of traffic forecasting due to their powerful spatial-temporal modeling capabilities. STGNNs are a class of prediction models built by integrating GNNs responsible for spatial modeling and temporal modules (e.g., Recurrent Neural Networks (RNNs) \cite{sutskever2014sequence,li2018dcrnn_traffic,bai2020adaptive,chen2021z,yu2022regularized,liu2022msdr}, Convolutinal Neural Networks (CNNs) \cite{yu2018spatio,wu2019graph,huang2020lsgcn,song2020spatial,li2021spatial}, and attention mechanisms \cite{guo2019attention,wang2020traffic,zheng2020gman,lan2022dstagnn,cirstea2022towards}).

\begin{figure}[t]
\centering
\includegraphics[width=.95\linewidth]{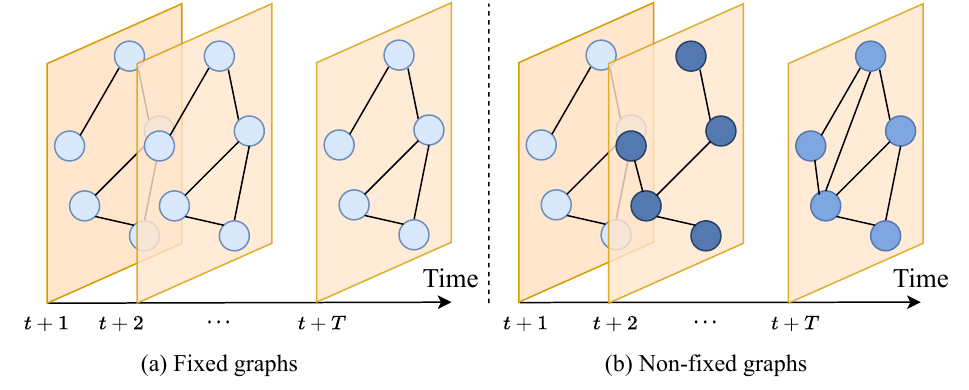} % Reduce the figure size so that it is slightly narrower than the column.
% \vspace{-0.1in}
\caption{Graph constructions. The underlying road network graphs at different time steps are not fixed but evolving.}
\label{Example1}
% \vspace{-0.1in}
\end{figure}

\IEEEpubidadjcol

\begin{figure*}[t]
\centering
\includegraphics[width=\linewidth]{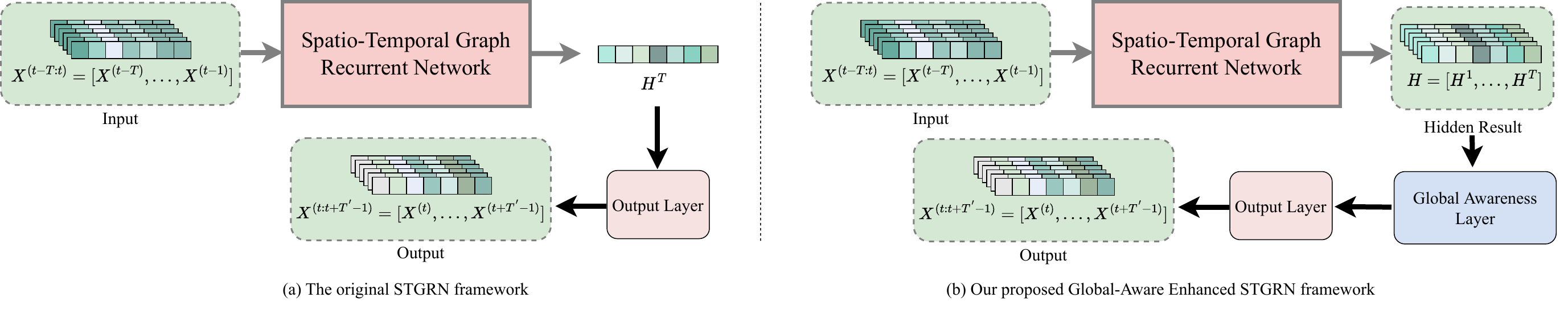} % Reduce the figure size so that it is slightly narrower than the column.
% \vspace{-0.1in}
\caption{Frameworks: STGRN and GA-STGRN.}
\label{Example2}
% \vspace{-0.1in}
\end{figure*}

However, there are still some issues with the spatial modeling and framework: (1) Static graphs or adaptive graphs overlook the dynamic changes in the graph structure as well as edge weights. When using a static graph for spatial modelling, one needs to pre-specify it in advance, thus having limited modeling power. While previous adaptive graph methods (e.g. AGCRN \cite{bai2020adaptive}) automatically learn the graph structure, all of them made the assumption that the graph structure is fixed and uniform at all time steps, as shown in Figure \ref{Example1}(a). In fact, this may not be true in real-world networks, as the road network graphs at different time steps are not fixed but evolving, as shown in Figure \ref{Example1}(b). (2) Spatial-temporal graph recurrent neural networks (STGRNs) lack the ability to perceive long-range temporal dependencies between different time steps as well as highly dynamic spatial features. STGRNs, a class of STGNNs comprising GNNs and RNNs such as AGCRN \cite{bai2020adaptive}, Z-GCNETs \cite{chen2021z} and RGSL \cite{yu2022regularized}. Its framework structure is given in Figure \ref{Example2}(a). Although STGRNs have excellent prediction performance, they mainly focus on enhancement on the spatial modeling GNNs, and the internal loop operation in RNNs makes them difficult to capture global temporal and spatial correlations in
traffic data.

To enhance the global information awareness of STGRNs, we propose a new framework: Global-Aware Enhanced Spatial-Temporal Graph Recurrent Networks (GA-STGRN), as shown in Figure \ref{Example2}(b). GA-STGRN essentially extends the original STGRN framework (in Figure \ref{Example2}(a)) by incorporating an additional global awareness layer. In our model construction, we systematically leverage attention mechanisms to design three distinct Global Spatial-Temporal Transformer-like architectures ($\mathrm{GST^2}$) specifically for the global awareness layer. These $\mathrm{GST^2}$ can seamlessly integrate into the existing STGRNs and improve their predictive performance. Remarkabaly, the enhancement is fairly general because STGRNs are a wide and typical class of methods including AGCRN, Z-GCNETs, TAMP-S2GCNets \cite{chen2021tamp}, RGSL, and ZFC-SHCN \cite{chen2022time}. Before this work, it is unknown how to systematically use spatial and temporal attentions to improve existing STGRNs. Meanwhile, we introduce a sequence-aware graph learning module to adaptively learn distinct graph structures over multiple time steps during training, which are more dynamic and powerful. A remarkable contribution of this work is to propose a new traffic prediction framework GA-STGRN to overcome the limitations of the STGRN framework. Last but not the least, our comprehensive experiments on 20 baselines and 4 datasets demonstrate the improvement with ablation study, visualization, memory analysis, and source code supported. Our specific contributions are as follows:

\begin{itemize}
\item We propose the First-ever framework to enhance existing spatial-temporal graph recurrent neural networks (STGRNs) with the \emph{global-aware module}, named GA-STGRN. This is a novel framework that combines GNNs, GRU, and a global awareness layer as the foundation for constructing new predictive models.
\item We propose a sequence-aware graph learning module, which uses node embedding endowed with time-indexed information to initialise non-fixed graphs and learn continuously during training.
\item In the global awareness layer, we use attention mechanisms across spatial and temporal features to design three different global spatial-temporal Transformer-like architectures ($\mathrm{GST^2}$), effectively enhancing the global-awareness capability of the STGRN framework.
\item To validate our approach, we conduct extensive experiments on four real-world traffic flow datasets. The results demonstrate the superiority of our proposed GA-STGRN framework over STGRN. Moreover, all three predictive models consistently outperform the baseline methods.
\end{itemize}

\section{Related Work}
\subsection{Traffic Flow Prediction}
\noindent Early traffic prediction methods solely focused on capturing temporal information, including Historical Average (HA), Vector Auto-Regressive (VAR) \cite{zivot2006vector}, Auto-Regressive Integrated Moving Average (ARIMA) \cite{lee1999application,williams2003modeling}, Support Vector Machines (SVR) \cite{drucker1996support,wu2004travel}. These methods have low accuracy because they can only consider the linear relationship between the nodes themselves. 

Recently, deep neural network models have received significant attention for their powerful predictive capabilities. Three commonly used networks for temporal modeling are RNNs \cite{sutskever2014sequence,li2018dcrnn_traffic,bai2020adaptive,chen2021z,yu2022regularized,liu2022msdr}, CNNs \cite{yu2018spatio,wu2019graph,huang2020lsgcn,song2020spatial,li2021spatial}, and attention mechanisms \cite{guo2019attention,lan2022dstagnn,cirstea2022towards}. For spatial modeling, graph neural networks (GNNs) \cite{defferrard2016convolutional,kipf2016semi} have emerged due to their ability to capture complex spatial relationships in unstructured spatial-temporal data. To boost predictive performance, researchers have been focusing on effectively combining GNNs with one or more temporal models to construct spatial-temporal graph neural networks. These networks can be categorized into RNN-based, CNN-based, and attention-based models from the perspective of temporal modeling. In some classic works, DCRNN \cite{li2018dcrnn_traffic} proposes diffusion graph convolution and integrates it with GRU for traffic prediction; STGCN \cite{yu2018spatio} combines 1D CNN and graph convolution to capture spatial-temporal correlation; ASTGCN \cite{guo2019attention} integrates spatial-temporal attention mechanism, graph convolution and temporal CNN to capture dynamic spatial-temporal features. Additionally, some works \cite{fang2021spatial,choi2022graph} incorporate differential equations into neural networks for traffic prediction, achieving commendable predictive accuracy.

\subsection{Graph Convolution Networks}
Graph convolution networks that do not rely on graph theory and define convolution operations directly in space are called spatial graph convolution. GNN 
 \cite{hechtlinger2017generalization} forcefully varies the data of a graph structure into a rule-like data, thus achieving 1-dimensional convolution. GraphSAGE \cite{hamilton2017inductive} performs convolution operations by sampling and information aggregation. GAT \cite{velivckovic2017graph} uses the attention mechanism to differentially aggregate neighboring nodes. And the graph convolution network that uses the graphical spectral theory and convolution theorem and transfers the data from the null domain to the spectral domain is called spectral domain graph convolution \cite{bruna2013spectral}. ChebNet \cite{defferrard2016convolutional} simplifies computation by using Chebyshev polynomials instead of convolution kernels in the spectral domain, specifically to reduce the eigenvalue decomposition of Laplacian matrices. GCN \cite{kipf2016semi} only considers Chebyshev polynomials of the 1st order.

\subsection{Attention Mechanisms}
Attention mechanisms are widely applied across various fields of artificial intelligence due to its powerful generalization and interpretability \cite{radford2019language,bai2021segatron,lin2021end}. The primary objective of the attention mechanism is to identify and prioritize crucial information relevant to the current task from a vast pool of available data. It draws inspiration from the selective visual attention mechanism in humans. Scaled Dot-Product Attention stands out as one of the extensively employed methods among various attention mechanisms. Transformer \cite{vaswani2017attention} abandons traditional architectures such as RNN or CNN and introduces attention mechanisms, achieving state-of-the-art performance in various domains. If we draw on the Transformer architecture for traffic flow prediction, we need to consider temporal and spatial modeling separately, and fusing temporal and spatial attention mechanisms is one way to do so.

\section{Proposed Methods}
\begin{figure}[t]
\centering
\includegraphics[width=\linewidth]{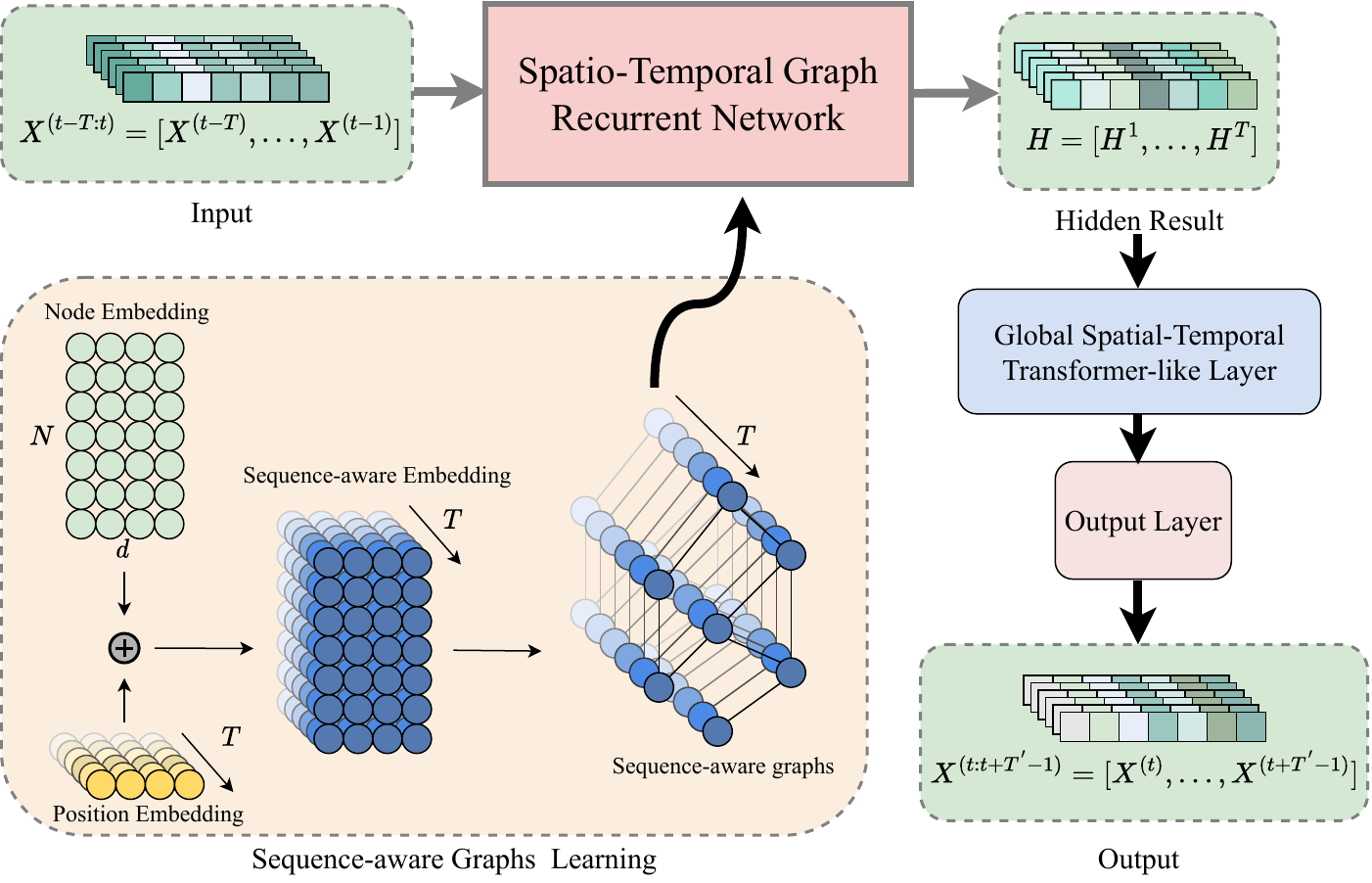} % Reduce the figure size so that it is slightly narrower than the column.
\caption{Model Overview of $\mathrm{GST^2}$-STGRN.}
\label{model}
% \vspace{-0.1in}
\end{figure}

In this section, we initially provide the mathematical definition for the problem of predicting traffic flow. We then develop new spatial-temporal neural networks based on the GA-STGRN framework, called Spatial-Temporal Graph Recurrent Neural Networks with Global Spatial-Temporal Transformer-like layer ($\mathrm{GST^2}$-STGRN), and describe technical details. Figure \ref{model} shows the overall structure of the $\mathrm{GST^2}$-STGRN.

\subsection{Problem Definition}
\noindent The traffic prediction task aims to make full use of the traffic road network $G$ and the complex spatial-temporal information in the historical traffic data to achieve accurate prediction of future traffic flow. where $V$ represents $N=|V|$ nodes in the traffic road network (e.g., observation points and road segments), $E$ is the set of edges, and $A\in {{\mathbb{R}}^{N \times N}}$ denotes the adjacency matrix of correlations between nodes. The traffic forecasting problem can be expressed as learning the forecasting function $F$ from the past $T$ steps of traffic flow $X^{(t-{T}:t)}=[X^{(t-T)}, \dots, X^{(t-1)}]$ and the road network graph $G$ to forecast the traffic flow $X^{(t:{t+{T}^{'}})}=[X^{(t)}, \dots, X^{(t+{T}^{'}-1)}]$ at the next ${T}^{'}$ steps:
\begin{equation}\label{eq1}
[X^{(t-T)}, \dots, X^{(t-1)}, G]\stackrel{F_{\Theta}}{\longrightarrow}[X^{(t)}, \dots, X^{(t+{T}^{'}-1)}],
\end{equation}
where $\Theta$  represents all the trainable parameters within the prediction function $F$.

\subsection{Sequence-Aware Graph Convolution Network (SGCN)}
\noindent The adjacent nodes of the traffic road network are highly correlated with the propagation of traffic flow, and there may be implicit relationships between non-adjacent nodes due to factors such as location and time period. Therefore,  optimizing the efficacy of spatial modeling emerges as a pivotal determinant of our model's predictive prowess. Our work uses ChebNet \cite{defferrard2016convolutional}—an archetypal spectral domain graph convolutional network—to facilitate intricate spatial modeling.
\begin{equation}\label{eq2}
Z = g_{\theta }\star_{G}x = \sum_{k=0}^{K}\theta_{k}T_{k}(\tilde{L})x,
\end{equation}
where $\beta$ is the learnable parameter and $K$ is the order. ChebNet uses Chebyshev polynomials: $T_{0}(\tilde{L}) = I, T_{1}(\tilde{L}) = \tilde{L}$, and $T_{n+1}(\tilde{L}) = 2\tilde{L}T_{n}(\tilde{L}) - T_{n-1}(\tilde{L})$. $\tilde{L} = \frac{2}{\lambda_{max}}L-I$ is the scaled Laplacian matrix, where $\lambda_{max}$ is the largest eigenvalue and $L=I-D^{-\frac{1}{2}}{A}D^{-\frac{1}{2}} \in \mathbb {R}^{N \times N}$ is the symmetric normalized graph Laplacian matrix.

Currently, multiple studies \cite{bai2020adaptive,yu2022regularized} focus on researching adaptive adjacency matrices and have achieved commendable outcomes. This is accomplished by initializing node embedding tensors and efficiently learning the adjacency matrix during the training process. Nevertheless, a prerequisite for its successful implementation is the assumption of consistent graph structure across all time steps, a condition that may not hold true in real-world scenarios. To this end, we propose a Sequence-aware Graph Learning module to generate different adjacency matrices for different time steps, which are relatively dynamic. We randomly initialize a learnable node embedding $E_n \in \mathbb {R}^{N \times d}$ and a Position embedding $E_p \in \mathbb {R}^{T \times 1 \times d}$, which respectively denote the static information of each node and the unique information relative to time, where $d$ denotes the embedding dimension. Initialize the adaptive scaled Laplacian matrix for time step $t$:
\begin{equation}
    \begin{split}\label{eq3}
    T_{1}(\tilde{L})[i] = \tilde{L}[i] &= softmax(E[i]\cdot E[i]^{\mathbf{T}}), \\
    \text{where}\ E[i]&=LayerNorm(E_n+E_p[i]),
    \end{split}
\end{equation}

To explore the hidden spatial correlations between nodes at different orders, we concatenate $T_{k}(\hat{L})$ under K-order ChebNet as a tensor $\tilde{T} = [I, T_{1}(\hat{L}), T_{2}(\hat{L}), \dots, T_{K}(\hat{L})]^{\mathbf{T}}\in \mathbb {R}^{(K+1) \times T \times N \times N}$. We use the independent node-parameter approach to convolution operations proposed by \cite{bai2020adaptive}. The sequence-aware graph convolution operation of the graph signal $X^{(i)}$ can be expressed as:
\begin{equation}\label{eq4}
SGCN(X^{(i)}) = {\tilde{T}[:,i]}X^{(i)}{E[i]}W_i+{E[i]}b_i,
\end{equation}
where $W_i \in \mathbb {R}^{d \times K \times d \times d}$ and $b_i \in \mathbb {R}^{d \times d}$ are the learnable parameters.

\begin{figure*}[t]
\centering
\includegraphics[width=\textwidth]{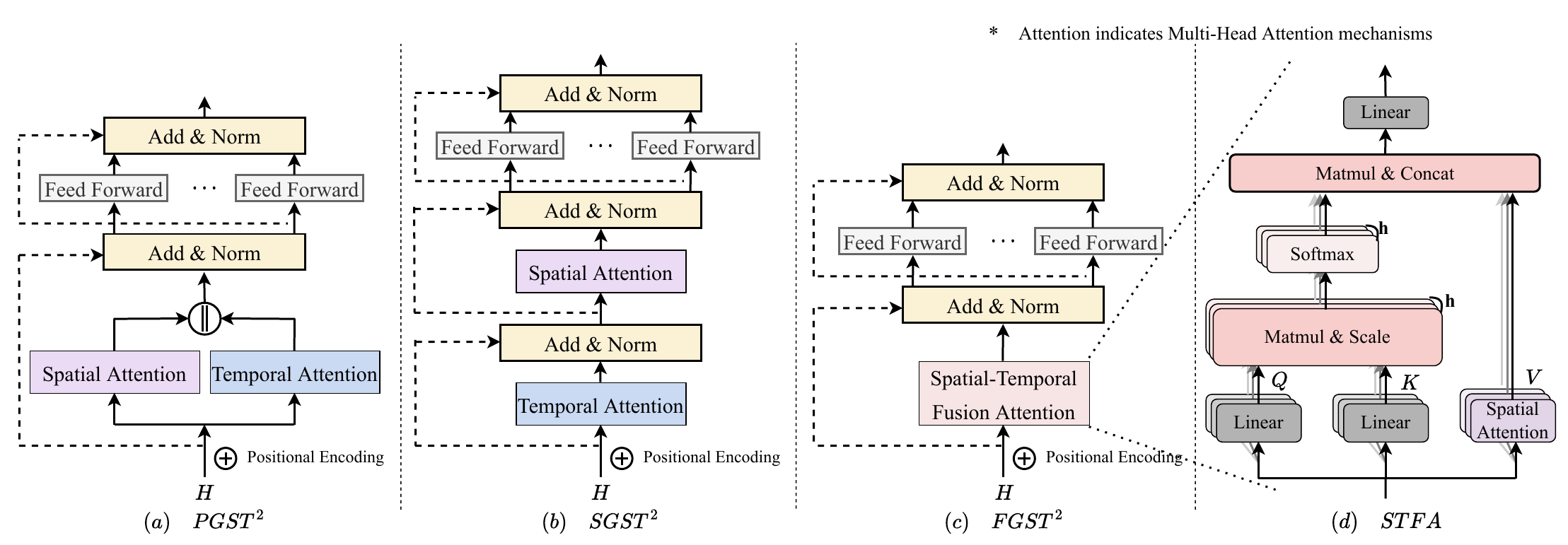} % Reduce the figure size so that it is slightly narrower than the column.
% \vspace{-0.2in}
\caption{Global Spatial-Temporal Transformer-like architectures ($\mathrm{GST^2}$). (a), (b) and (c) are three different ways of constructing $\mathrm{GST^2}$, and (d) is a component of $\mathrm{FGST^2}$: Spatial-Temporal Fusion Attention (STFA).}
\label{att}
% \vspace{-0.1in}
\end{figure*}

\subsection{Spatial-Temporal Graph Recurrent Network}
To further analyse the unique temporal and spatial dependencies between traffic flow sequences, we integrate the SGCN into the GRU to construct a spatial-temporal graph recurrent network. The specific equation can be expressed as follows:
\begin{equation}
    \begin{split}\label{eq5}
    z^{(i)} &= \sigma(SGCN([{H_{m}}^{(i)},h^{(i-1)}])), \\
    r^{(i)} &= \sigma(SGCN([{H_{m}}^{(i)},h^{(i-1})])), \\
    \tilde{h}^{(i)} &= tanh(SGCN([{H_{m}}^{(i)},r^{(i)} \odot h^{(i-1)}])), \\
    h^{(i)} &= z^{(i)}\odot h^{(i-1)} + (1-z^{(i)})\odot \tilde{h}^{(i)},
    \end{split}
\end{equation}
where we employ two activation functions—specifically,  the Sigmoid function denoted as $\sigma$ and the Tanh function denoted as $\tanh$.

\subsection{Global Spatial-Temporal Transformer-like Layer}
Although the STGRN framework is proficient in capturing sequential dependencies, its long-term prediction performance is constrained by its structural features. In order to fully capture the long-term temporal dependence and complex spatial dependence, we design the Global Spatial-Temporal Transformer-like architecture ($\mathrm{GST^2}$) as the global awareness layer.

Self-attention is effective in capturing long-term dependencies and is used in the temporal or spatial modeling of some works \cite{guo2019attention,lan2022dstagnn}. Here, we choose to scale dot product attention, which is one of the self-attention mechanisms and is formulated as follows:
\begin{equation}\label{eq6}
Q=XW_{q}, K=XW_{k}, V=XW_{v},
\end{equation}    
\begin{equation}\label{eq7}
Att(Q, K, V) = softmax(\frac{QK^{\mathbf{T}}}{\sqrt{d}})V,
\end{equation}
where query $Q$, key $K$ and value $V$ come from the same input signal $X$ and $d$ is the dimension of $Q,K$ and $V$. Temporal attention (TA) and spatial attention (SA) are important components of $\mathrm{GST^2}$: (1) {\bf{Temporal attention}} is to capture the correlation present in traffic conditions between different time steps in the temporal dimension. In order to learn the dependencies in different modes in parallel, we use multi-headed attention. The multi-headed temporal attention formula is as follows:
\begin{equation}
    \begin{split}\label{eq8}
    TA(X) &= Concat(head_{1}, \dots, head_{h})W_o, \\
    \text{where}\ head_{j} &= Att(XW^j_q, XW^j_k, XW^j_v),
    \end{split}
\end{equation}
$X \in \mathbb {R}^{N \times T \times d}$ is the input, $h$ is the number of heads, and the learnable parameters $W_o \in \mathbb {R}^{d \times d}$, $W^j_{q},W^j_{k},W^j_{v} \in \mathbb {R}^{d \times d/h}$. 

(2) {\bf{Spatial attention}} captures the spatial dependencies that exist between different traffic detectors in the spatial dimension. It is similar to the temporal attention structure but with different dimensions of attention weights. The dimension of $X$ for temporal attention is $\mathbb {R}^{N \times T \times d}$, where $N$ is the number of nodes and $T$ is the time dimension, while for spatial attention $SA(X)$, $X$ needs to be reshaped as $\mathbb {R}^{T \times N \times d}$. For the convenience of the following description, our specific procedure for $SA(X)$ is to first reshape $X \in \mathbb {R}^{N \times T \times d}$ to $\mathbb {R}^{T \times N \times d}$,  then perform the attention calculation, and finally output after reshaping to $\mathbb {R}^{T \times N \times d}$. 

When taking the output from the spatial-temporal graph convolution layer to the global spatial-temporal Transformer-like layer, it is first merged with the positional encoding to make better use of the relative order information. The positional encoding is calculated as follows:
\begin{equation}
    \begin{split}\label{eq9}
    PE(t, 2c) &= sin(t/1000^{{2c}/{d}}), \\
    PE(t, 2c+1) &= cos(t/1000^{{2c}/{d}}),
    \end{split}
\end{equation}
where $t$ is the relative position index. In order to capture long-term temporal and spatial correlations from traffic sequences, we choose temporal attention and spatial attention, but how to effectively combine TA and SA to construct a more effective $\mathrm{GST^2}$? Here we systematically study 3 effective combination methods of spatial and temporal attentions and construct three different $\mathrm{GST^2}$.
%the Parallel Global Spatial-Temporal Transformer-like architecture  ($\mathrm{PGST^2}$), the Serial Global Spatial-Temporal Transformer-like architecture ($\mathrm{SGST^2}$), and the Fused Global Spatial-Temporal Transformer-like architecture ($\mathrm{FGST^2}$).

\subsubsection{Parallel $\mathrm{GST^2}$}
The calculation of TA and SA in $\mathrm{PGST^2}$ is parallel with the same input data. The specific operational procedure is to first input $H \in \mathbb {R}^{N \times T \times d}$ and positional encoding to integrate to obtain $H_e$, then $H_e$ is used as input to both SA and TA, and finally the combined result $H_p$ is input to the fully connection network. Residual connection and layer normalisation are added after both the attention operation and the fully connection network, as are the other two $\mathrm{GST^2}$. The specific model implementation is shown in Figure \ref{att}(a), and the formula is as follows:
\begin{equation}
    \begin{split}\label{eq10}
    H_e &= H+PE, \\
    H_c &= Concat(TA(H_e), SA(H_e)),\\
    H_p &= LayerNorm(H_c+H_e), \\
    H_o &= LayerNorm(FC(H_p)+H_p). \\
    \end{split}
\end{equation}

\subsubsection{Serial $\mathrm{GST^2}$}
Unlike $\mathrm{PGST^2}$, the computation of TA and SA in $\mathrm{SGST^2}$ is sequential, i.e. the structures are connected in series. The specific operation process is to first input $H$ and positional encoding integration to obtain $H_e$. Then $H_e$ is used as input to the attention mechanism, which goes through TA and SA sequentially, and finally the result $H_s$ is input to the fully connected network. The specific model implementation is shown in Figure \ref{att}(b), and the formula is as follows:
\begin{equation}
    \begin{split}\label{eq11}
    H_e &= H+PE, \\
    H_t &= LayerNorm(TA(H_e)+H_e),\\
    H_s &= LayerNorm(SA(H_t)+H_t), \\
    H_o &= LayerNorm(FC(H_s)+H_s). \\
    \end{split}
\end{equation}

\subsubsection{Fused $\mathrm{GST^2}$}
TA and SA capture temporal and spatial dependencies separately, so we want to integrate them effectively in the process of exploring their combination methods to achieve simultaneous capture of temporal and spatial dependencies. To this end, we propose a new attention mechanism: {\bf{Spatial-temporal fusion attention}} (STFA). The specific model implementation is shown in Figure \ref{att}(d), and the formula is as follows:
\begin{equation}
    \begin{split}\label{eq12}
    STFA(X) &= Concat(head_{1}, \dots, head_{h})W_o, \\
    \text{where}\ head_{j} &= Att(XW^j_q, XW^j_k, SA(X)).
    \end{split}
\end{equation}

The specific operation flow is to first input $H$ and positional encoding to obtain $H_e$, then $H_e$ is used as input to STFA, and finally the output result $H_f$ is input to the fully connected network. The specific model implementation is shown in Figure \ref{att}(c), and the formula is as follows:
\begin{equation}
    \begin{split}\label{eq13}
    H_e &= H+PE, \\
    H_f &= LayerNorm(STFA(H_e)+H_e), \\
    H_o &= LayerNorm(FC(H_f)+H_f). \\
    \end{split}
\end{equation}

\subsection{Output Layer}
After $\mathrm{GST^2}$, we take the result $H_o$ as input and use a two-layer fully connected network with a ReLU activation in between to predict the traffic flow at future $T$ time steps:
\begin{equation}\label{eq14}
X^{(t:{t+{T}^{'}})} = FC(Relu(FC({H_o})))),
\end{equation}

We use L1 loss to train the objective function:
\begin{equation}\label{eq15}
Loss = \frac{1}{{T}^{'}} ({\hat{X}}^{(t:{t+{T}^{'}})}-X^{(t:{t+{T}^{'}})}),
\end{equation}
where ${\hat{X}}^{(t:{t+{T}^{'}})}$ is the ground-truth traffic data.

\section{Experimental Results}
% In this section, we demonstrate the validity of DMSTRN through a series of experiments. We first describe the datasets and the experimental setup, and then analyse the predicted results obtained. Finally, we provide a comprehensive discussion of the ablation study and hyperparameter tuning.

\subsection{Datasets}
We perform experiments with four real-world traffic datasets sourced from the Caltrans Performance Measure System (PeMS) \cite{chen2001freeway}, specifically PEMSD3, PEMSD4, PEMSD7, and PEMSD8 \cite{fang2021spatial}. The traffic data is collected at 5-minute intervals, generating a total of 288 data points per day. In our experiments, we only use traffic flow data. Details of these datasets are provided in Table \ref{table1}.
\begin{table}[t]
\centering
\caption{Statistics of the tested datasets}
	\label{table1}
\resizebox{.85\columnwidth}{!}{
	\begin{tabular}{c c c c c c}
		\toprule[1.2pt]
		Datasets & Nodes & Samples & Unit & Time Span \\
		\midrule
		PEMSD3 & $358$ & $26,208$ & $5$ mins & $3$ months \\
		PEMSD4 & $307$ & $16,992$ & $5$ mins & $2$ months \\
		PEMSD7 & $883$ & $28,224$ & $5$ mins & $4$ months \\
		PEMSD8 & $170$ & $17,856$ & $5$ mins & $2$ months \\
		\bottomrule[1.2pt]
	\end{tabular}
	}
        % \vspace{-0.1in}
	% \caption{Statistics of the tested datasets}
	% \label{table1}
  % \vspace{-0.1in}
\end{table}

\subsection{Baseline Methods}
We compared our models with the following 20 baseline methods and categorised them as:
% \begin{itemize}
%     \item Traditional time series forecasting methods, Historical Average (HA), ARIMA \cite{williams2003modeling}, VAR \cite{zivot2006vector}, and SVR \cite{drucker1996support}; 
%     \item RNN-based models: FC-LSTM \cite{sutskever2014sequence}, DCRNN \cite{li2018dcrnn_traffic}, AGCRN \cite{bai2020adaptive}, Z-GCNETs \cite{chen2021z}, RGSL \cite{yu2022regularized}, and  GMSDR \cite{liu2022msdr};
%     \item CNN-based methods: STGCN \cite{yu2018spatio}, Graph WaveNet \cite{wu2019graph}, LSGCN \cite{huang2020lsgcn}, STSGCN \cite{song2020spatial}, and STFGNN \cite{li2021spatial}; 
%     \item Attention-based models: ASTGCN(r) \cite{guo2019attention}, DSTAGNN \cite{lan2022dstagnn}, and  ST-WA \cite{cirstea2022towards}; 
%     \item Differential equation-based methods: STGODE \cite{fang2021spatial} and STG-NCDE \cite{choi2022graph}. 
% \end{itemize}
\begin{itemize}
    \item Traditional time series forecasting methods: (1) \textbf{HA}, which uses the average value of historical traffic flows to predict future traffic flows; (2) \textbf{ARIMA} \cite{williams2003modeling}, which is a widely used model for time series forecasting; (3) \textbf{VAR} \cite{zivot2006vector}, a statistical model that captures the relationship between multiple variables over time and (4) \textbf{SVR} \cite{drucker1996support}, which uses linear support vector machines for regression tasks. 
    \item RNN-based models: (5) \textbf{FC-LSTM} \cite{sutskever2014sequence}, LSTM network with fully connected hidden units; (6) \textbf{DCRNN} \cite{li2018dcrnn_traffic}, which designed to capture spatial dependencies through diffuse graph convolution and temporal dependencies through an encoder-decoder network architecture; (7) \textbf{AGCRN} \cite{bai2020adaptive}, which integrates adaptive graph generation and node adaptive parameter learning with traditional graph convolution in a recurrent neural network, enhancing its ability to capture complex spatial-temporal correlations; (8) \textbf{Z-GCNETs} \cite{chen2021z}, which introduces the concept of Zigzag persistence to time-aware graph convolutional networks; (9) \textbf{RGSL} \cite{yu2022regularized}, which incorporates a Regularized Graph Generation module for learning implicit graphs, followed by the Laplacian Matrix Mixed-up module to integrate explicit and implicit structures and (10) \textbf{GMSDR} \cite{liu2022msdr}, integrating graph neural networks with Multi-Step Dependency Relation, a new variant of recurrent neural networks, for spatial-temporal prediction.
    \item CNN-based methods: (11) \textbf{STGCN} \cite{yu2018spatio}, which combines graph convolution and 1D convolution to capture spatial-temporal correlations; (12) \textbf{Graph WaveNet} \cite{wu2019graph}, which combines diffuse graph convolution with 1D convolution while introducing an adaptive adjacency matrix; (13) \textbf{LSGCN} \cite{huang2020lsgcn}, which integrates a novel graph attention network with graph convolution within a spatial gated block; (14) \textbf{STSGCN} \cite{song2020spatial}, which efficiently extracts localized spatial-temporal correlations by incorporating a carefully designed local spatial-temporal subgraph module and (15) \textbf{STFGNN} \cite{li2021spatial}, which incorporates a newly designed spatial-temporal fusion graph module, parallelly integrated with a 1D convolution module.
    \item Attention-based models: (16) \textbf{ASTGCN(r)} \cite{guo2019attention},  which integrates spatial and temporal attention mechanisms with spatial-temporal convolution to effectively capture dynamic spatial-temporal features; (17) \textbf{DSTAGNN} \cite{lan2022dstagnn}, which introduces a novel dynamic spatial-temporal awareness graph, replacing the conventional static graph employed by traditional graph convolution methods; and (18) \textbf{ST-WA} \cite{cirstea2022towards}, which consists of multiple layers of Spatial-Temporal Aware Window Attention.
    \item Differential equation-based methods: (19) \textbf{STGODE} \cite{fang2021spatial}, which employs a tensor-based ordinary differential equation (ODE) to effectively capture spatial-temporal dynamics and (20) \textbf{STG-NCDE} \cite{choi2022graph}, which introduces two NCDEs tailored for temporal and spatial processing, seamlessly integrating them into a unified framework. 
\end{itemize}

\subsection{Experimental Settings}
Our proposed models were implemented in the PyTorch framework and all experiments were conducted on an NVIDIA GTX 1080 TI GPU with 11GB of memory.
All four traffic datasets were divided into training, validation and test sets with ratio $6:2:2$. The experiments use traffic data from the last $12$ continuous time steps ($60$ minutes) to predict traffic flow for the next $12$ continuous time steps. In the training phase of the model we use the Adam optimizer with a learning rate set to $0.003$. The main model hyperparameters include the node embedding size $d \in {\{1, 2, \cdots, 10\}}$, the Chebyshev graph convolution order $K \in {\{1, 2, 3\}}$, and the weight decay coefficient $w \in {\{0, 0.0001, \cdots, 0.001\}}$. In addition, to prevent overfitting, we use Dropout operations for the input signal of $\mathrm{GST^2}$ and in the structure with decay probabilities $\alpha, \beta \in {\{0, 0.1\}}$.  In addition, we set the epochs to $500$ and use an early stopping strategy with a patience value of $30$. 

\begin{figure}[t]
\centering
\includegraphics[width=\columnwidth]{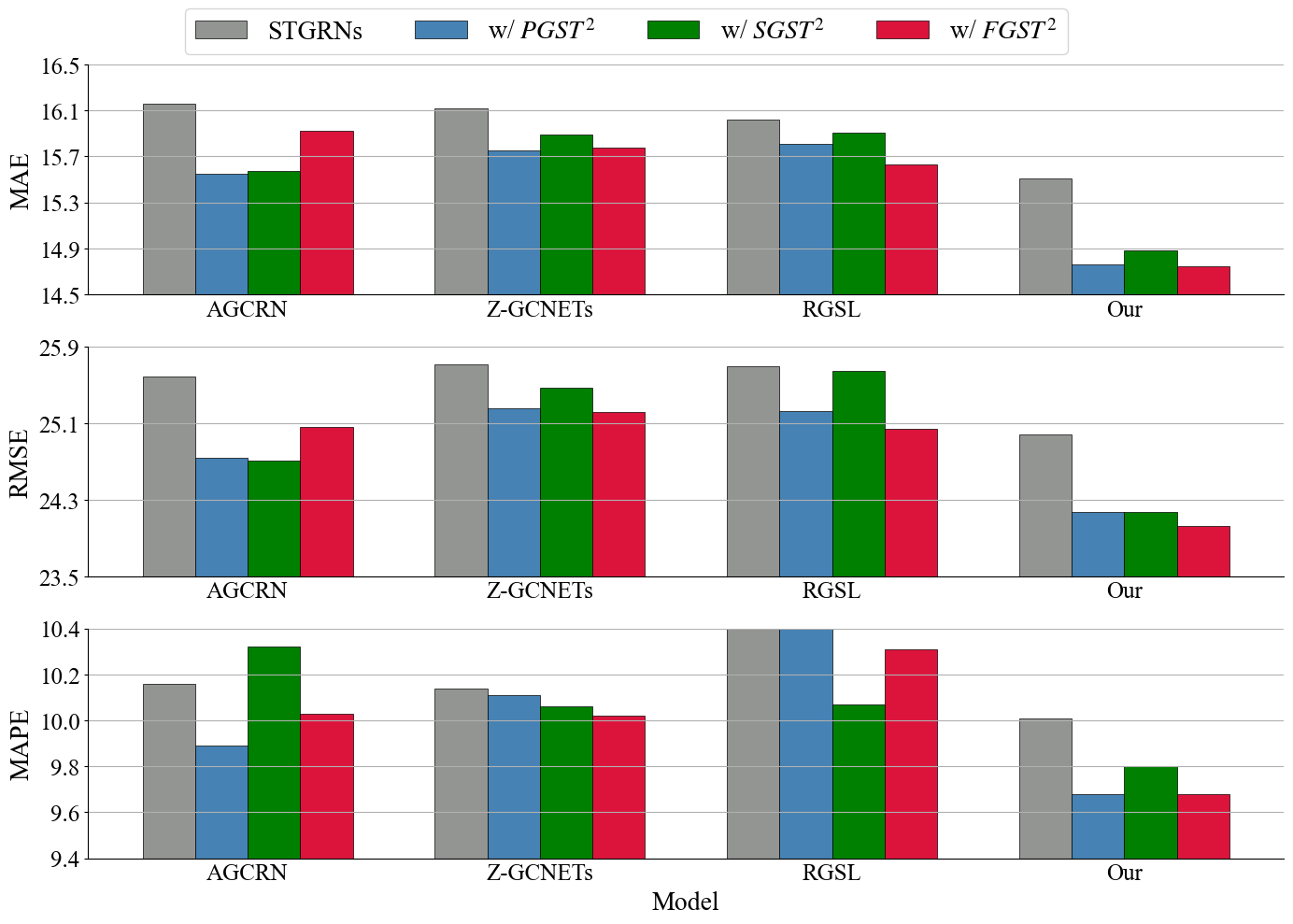}
% \vspace{-0.25in}
\caption{Applying our new framework in typical STGRNs}
\label{fram}
% \vspace{-0.2in}
\end{figure}

To assess the performance of all models, we employ three standard metrics:To assess the performance of all models, we employ three standard metrics: Mean Absolute Error (\emph{MAE}), Root Mean Square Error (\emph{RMSE}), Mean Absolute Percent Error (\emph{MAPE}). 
% Here are their formal definitions:
% \begin{equation}
% \begin{aligned}
% &\operatorname{MAE}(\hat{Y}, Y)=\frac{1}{T} \sum_{i=1}^{T}\left|\hat{y}_{i}-y_{i}\right| \\
% &\operatorname{RMSE}(\hat{Y}, Y)=\sqrt{\frac{1}{T} \sum_{i=1}^{T}\left(\hat{y}_{i}-y_{i}\right)^{2}} \\
% &\operatorname{MAPE}(\hat{Y}, Y)=\frac{100\%}{T} \sum_{i=1}^{T}\left|\frac{\hat{y}_{i}-y_{i}}{\hat{y}_{i}}\right|
% \end{aligned}
% \end{equation}
% where $\hat{Y}=\hat{y}_{1}, \hat{y}_{2}, \dots, \hat{y}_{T}$ is the real traffic data, $Y={y}_{1}, {y}_{2}, \dots, {y}_{T}$ is the predicted data, and $T$ is the predicted time step. In our experiments, $T=12$.
In the following discussion, we refer to the specific $\mathrm{GST^2}$-STGRN models using $\mathrm{PGST^2}$, $\mathrm{SGST^2}$ and $\mathrm{FGST^2}$ in the global awareness layer as $\mathrm{PGST^2}$-STGRN, $\mathrm{SGST^2}$-STGRN and $\mathrm{FGST^2}$-STGRN, respectively.

\subsection{Experimental Results}
The main experimental results are divided into two parts: the advantages of the GA-STGRN framework and the performance comparison for the concrete models.

\begin{figure}[!t]
\centering
\subfloat[]{\includegraphics[width=1.7in]{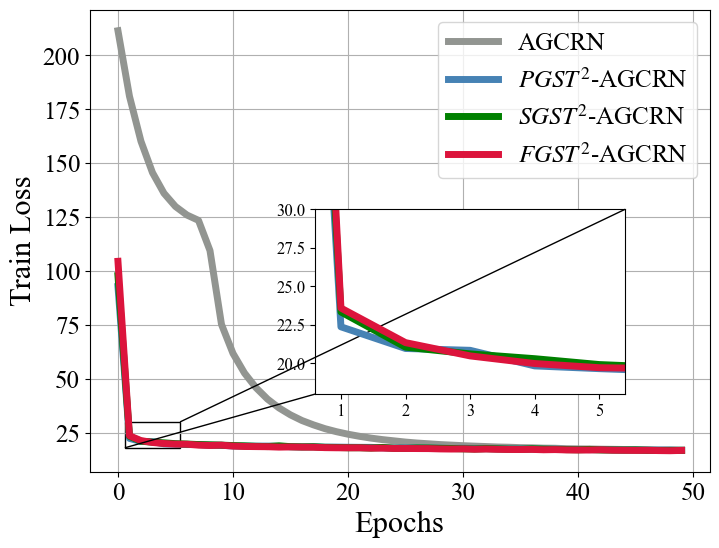}%
\label{Visual1}}
\hfil
\subfloat[]{\includegraphics[width=1.7in]{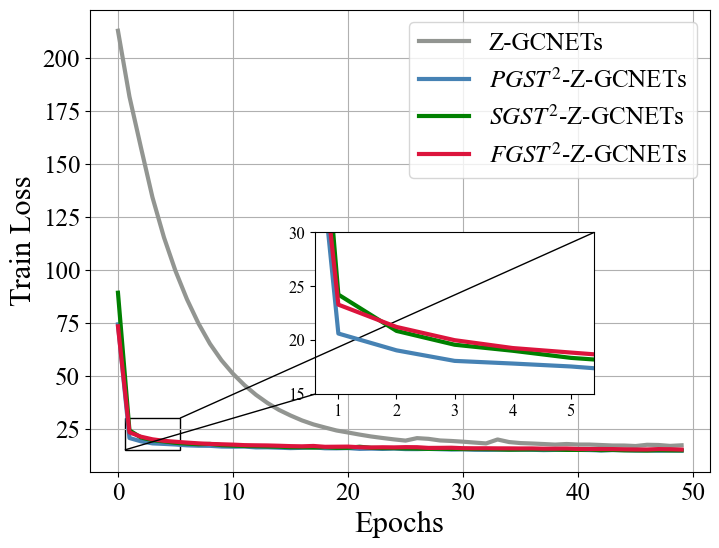}%
\label{Visual2}}

\subfloat[]{\includegraphics[width=1.7in]{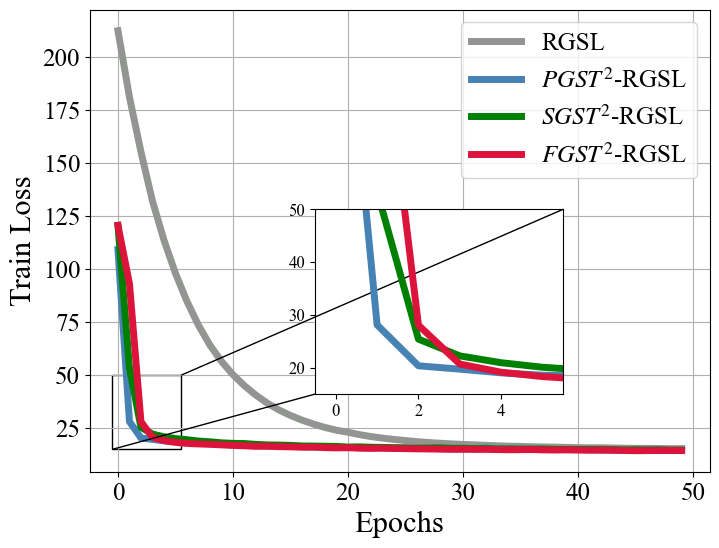}%
\label{Visual3}}
\hfil
\subfloat[]{\includegraphics[width=1.7in]{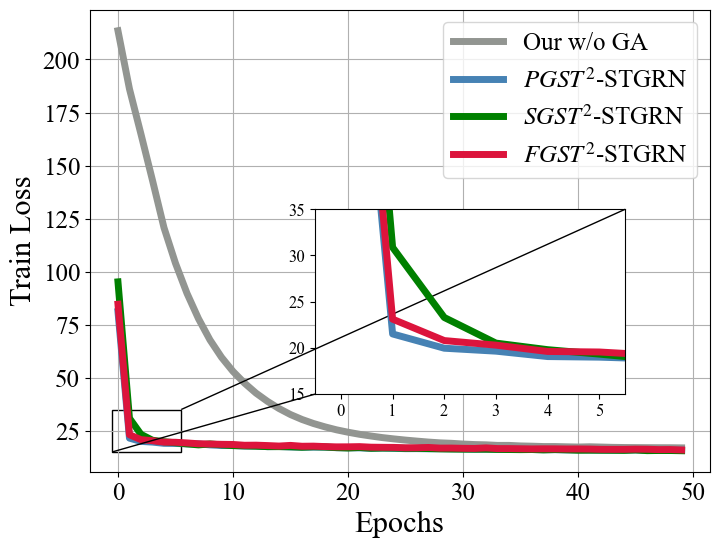}%
\label{Visual4}}
\caption{Training loss convergence speed.}
\label{fig_visual}
\end{figure}

\begin{table*}[t]
	\renewcommand{\arraystretch}{1.2}
	\centering
 \caption{Performance comparison of different models on the tested datasets. Our three models consistently outperform all the baseline methods, as shown in bold font. The underlined results are the current state-of-the-art results of the existing methods and the best prediction results of our proposed models, respectively.}
	\label{pred}
	\resizebox{\linewidth}{!}{
	\begin{tabular}{c|c c c|c c c |c c c|c c c}
		\toprule[1.2pt]
		\multirow{2}{*}{Model} & \multicolumn{3}{c|}{PEMSD3} & \multicolumn{3}{c|}{PEMSD4} & \multicolumn{3}{c|}{PEMSD7} & \multicolumn{3}{c}{PEMSD8} \\
		\cline{2-13}
		{}& MAE & RMSE & MAPE & MAE & RMSE & MAPE & MAE & RMSE & MAPE & MAE & RMSE & MAPE \\
		\midrule
		HA & 31.58 & 52.39 & 33.78\% & 38.03 & 59.24 & 27.88\% & 45.12 & 65.64 & 24.51\% & 34.86 & 59.24 & 27.88\% \\
		ARIMA & 35.41 & 47.59 & 33.78\% & 33.73 & 48.80 & 24.18\% & 38.17 & 59.27 & 19.46\% & 31.09 & 44.32 & 22.73\% \\
		VAR & 23.65 & 38.26 & 24.51\% & 24.54 & 38.61 & 17.24\% & 50.22 & 75.63 & 32.22\% & 19.19 & 29.81 & 13.10\% \\
		SVR & 20.73 & 34.97 & 20.63\% & 27.23 & 41.82 & 18.95\% & 32.49 & 44.54 & 19.20\% & 22.00 & 33.85 & 14.23\% \\ \hline
		FC-LSTM & 21.33 & 35.11 & 23.33\% & 26.77 & 40.65 & 18.23\% & 29.98 & 45.94 & 13.20\% & 23.09 & 35.17 & 14.99\%\\
		DCRNN(2018) & 17.99 & 30.31 & 18.34\% & 21.22 & 33.44 & 14.17\% & 25.22 & 38.61 & 11.82\% & 16.82 & 26.36 & 10.92\% \\
		AGCRN(2020) &15.98 & 28.25 & 15.23\% & 19.83 & 32.26 & 12.97\% & 22.37 & 36.55 & 9.12\% & 15.95 & 25.22 & 10.09\% \\ 
		Z-GCNETs(2021) & 16.64 & 28.15 & 16.39\% & 19.50 & 31.61 & 12.78\% & 21.77 & 35.17 & 9.25\% & 15.76 & 25.11 & 10.01\% \\ 
		RGSL(2022) & 15.85 & 28.51 & \underline{14.68\%} & 19.19 & 31.14 & 12.69\% & 20.58 & 33.88 & \underline{8.69\%} & 15.49 & 24.80 & 9.96\% \\
		GMSDR(2022) & 15.78 & 26.82 & 15.33\% & 20.37 & 32.52 & 13.71\% & 21.89 & 35.46 & 9.42\% & 16.36 & 25.58 & 10.28\% \\ \hline
		STGCN(2018) & 17.55 & 30.42 & 17.34\% & 21.16 & 34.89 & 13.83\% & 25.33 & 39.34 & 11.21\% & 17.50 & 27.09 & 11.29\% \\ 
		Graph WaveNet(2019) & 19.12 & 32.77 & 18.89\% & 24.89 & 39.66 & 17.29\% & 26.39 & 41.50 & 11.97\% & 18.28 & 30.05 & 12.15\% \\
		LSGCN(2020) & 17.94 & 29.85 & 16.98\% & 21.53 & 33.86 & 13.18\% & 27.31 & 41.46 & 11.98\% & 17.73 & 26.76 & 11.20\% \\
		STSGCN(2020) & 17.48 & 29.21 & 16.78\% & 21.19 & 33.65 & 13.90\% & 24.26 & 39.03 & 10.21\% & 17.13 & 26.80 & 10.96\% \\
		STFGNN(2021) & 16.77 & 28.34 & 16.30\% & 20.48 & 32.51 & 16.77\% & 23.46 & 36.60 & 9.21\% & 16.94 & 26.25 & 10.60\% \\ \hline
		ASTGCN(r)(2019) & 17.34 & 29.56 & 17.21\% & 22.93 & 35.22 & 16.56\% & 24.01 & 37.87 & 10.73\% & 18.25 & 28.06 & 11.64\% \\
		DSTAGNN(2022) &  15.57 & 27.21 &  \underline{14.68\%} & 19.30 & 31.46 &  12.70\% & 21.42 & 34.51 & 9.01\% & 15.67 &  24.77 & 9.94\% \\ 
		ST-WA(2022) & \underline{15.17} & \underline{26.63} & 15.83\% & \underline{19.06} & \underline{31.02} & \underline{12.52\%} & 20.74 & 34.05 & 8.77\% & \underline{15.41} & \underline{24.62} & 9.94\%\\ \hline
		STGODE(2021) & 16.50 & 27.84 & 16.69\% & 20.84 & 32.82 & 13.77\% & 22.59 & 37.54 & 10.14\% & 16.81 & 25.97 & 10.62\% \\
		STG-NCDE(2022) &  15.57 &  27.09 & 15.06\% &  19.21 &  31.09 & 12.76\% &  \underline{20.53} &  \underline{33.84} &  8.80\% &  15.45 & 24.81 &  \underline{9.92\%}\\
		\midrule
		$\mathrm{\bf{PGST^2}}${\bf{-STGRN}} & \bf{14.75} & \underline{\bf{25.57}} & \bf{14.12}\% & \bf{18.69} & \bf{30.72} & \bf{12.11}\% & \bf{19.99} & \bf{33.29} & \underline{\bf{8.41}} & \bf{14.78} & \bf{24.08} & \underline{\bf{9.51}\%} \\
		Improvements & +2.77\% & +3.98\% & +3.91\% & +1.94\% & +0.97\% & +3.27\% & +2.63\% & +1.63\% & +3.22\% & +4.09\% & +2.19\% & +4.13\% \\
        $\mathrm{\bf{SGST^2}}${\bf{-STGRN}} & \bf{14.77} & \bf{25.90} & \underline{\bf{13.72}\%} & \bf{18.71} & \underline{\bf{30.64}} & \bf{12.33}\% & \underline{\bf{19.91}} & \bf{33.34} & \underline{\bf{8.41}} & \underline{\bf{14.67}} & \underline{\bf{23.92}} & \bf{9.59}\% \\
		Improvements & +2.44\% & +2.74\% & +6.54\% & +1.84\% & +1.23\% & +1.52\% & +3.11\% & +1.48\% & +3.22\% & +4.80\% & +2.84\% & +3.33\% \\
        $\mathrm{\bf{FGST^2}}${\bf{-STGRN}} & \underline{\bf{14.74}} & \bf{26.01} & \bf{13.91}\% & \underline{\bf{18.62}} & \bf{30.72} & \underline{\bf{12.09}\%} & \bf{19.94} & \underline{\bf{33.21}} & \bf{8.42} & \bf{14.70} & \bf{23.97} & \bf{9.64}\% \\
		Improvements & +2.83\% & +2.33\% & +5.25\% & +2.31\% & +0.97\% & +3.43\% & +2.87\% & +1.86\% & +3.11\% & +4.61\% & +2.64\% & +2.91\% \\
		\bottomrule[1.2pt]
	\end{tabular}
	}
        % \vspace{-0.1in}
	% \caption{Performance comparison of different models on the tested datasets. Our three models consistently outperform all the baseline methods, as shown in bold font. The underlined results are the current state-of-the-art results of the existing methods and the best prediction results of our proposed models, respectively.}
	% \label{pred}
 % \vspace{-0.20in}
\end{table*}

% \smallskip
\subsubsection{Framework enhancement}
% {\noindent \bf Framework advantage:}
\noindent We consider several typical STGRNs, including AGCRN, Z-GCNETs and RGSL, feed them into our framework with the global awareness layer (Figure \ref{Example2}), and study how the proposed framework can boost their prediction performance. The original STGRNs integrate $\mathrm{PGST^2}$, $\mathrm{SGST^2}$ or $\mathrm{FGST^2}$ to form global aware enhanced STGRNs, which are expressed as w/ $\mathrm{PGST^2}$, w/ $\mathrm{SGST^2}$ or w/ $\mathrm{FGST^2}$, respectively. Figure \ref{fram} shows that the prediction performance of the models enhanced with the innovative $\mathrm{GST^2}$s are consistently better than the original STGRNs on PEMSD8. Additionally, our STGRN is more powerful than other STGRNs, due to the development of our sequence-aware graph learning module. We did not observe a clear winner among the 3 $\mathrm{GST^2}$ and thus would suggest trying all of them when possible.

Adding $\mathrm{GST^2}$ to STGRNs seems to naturally make the models more complex and increase the computation, but the experimental results show that their convergence speed becomes remarkably fast, which results in an even faster training time (to be discussed later in Table \ref{table_cost}). Figure 6 visualises the difference in the convergence speed of the model training error before and after global aware enhancement of the STGRNs. Ordinary STGRNs reach convergence at around the 30th epoch, while global aware enhanced STGRNs converge at the 5th epoch, which is 6 times faster than the former.

\begin{table}[t]
	\renewcommand{\arraystretch}{1.2}
	\centering
 \caption{Computation time and memory costs on PEMSD4.}
	\label{table_cost}
	\resizebox{\columnwidth}{!}{
	\begin{tabular}{c|c c c}
		\toprule[1.2pt]
        Model &  Training(s/epoch) & Inference(s/epoch) & Memory(MB) \\
		\midrule
		STGODE(2021) & 111.77 & 12.19 & 8773 \\
		Z-GCNETs(2021) & 63.34 & 7.40 & 8597 \\
		DSTAGNN(2022) &  242.57 & 14.64 & 10347 \\
		STG-NCDE(2022) & 1318.35 & 93.77 & 6091 \\
		RGSL(2022) & 99.36 & 20.51 & 6843 \\
		GMSDR(2022) & 125.19 & 19.20 & 5751 \\
		ST-WA(2022) & 84.43 & 4.92 & 4305 \\
		\midrule
		$\mathrm{\bf{PGST^2}}${\bf{-STGRN}} & 67.71 & 8.36 & 10868 \\
        $\mathrm{\bf{SGST^2}}${\bf{-STGRN}} & 66.08 & 8.11 & 10640 \\
        $\mathrm{\bf{FGST^2}}${\bf{-STGRN}} & 67.90 & 8.19 & 10538 \\
		\bottomrule[1.2pt]
	\end{tabular}
	}
        % \vspace{-0.1in}
	% \caption{Computation time and memory costs on PEMSD4.}
	% \label{table_cost}
 % \vspace{-0.2in}
\end{table} 

% \smallskip
\subsubsection{Model prediction performance}
% {\noindent \bf Model prediction performance:}
\noindent We now present the performance of our three concrete models compared to twenty (20) baseline methods on all four (4) test datasets. As shown in Table \ref{pred}, our three models outperform all baseline methods on all the datasets with significant performance boosting. In particular, on the PEMSD3 and PEMSD8 datasets, our models greatly improve on the current state-of-the-art methods by major margins. On the PEMSD3 dataset, the MAE, RMSE and MAPE values are $2.83\%$, $3.98\%$ and $6.54\%$ higher than the state-of-the-art method, respectively. Similarly, on the PEMSD8 dataset, these three values are improved by $4.80\%$, $2.84\%$ and $4.13\%$ respectively.

\begin{figure*}[!t]
\centering
\subfloat[]{\includegraphics[width=2.2in]{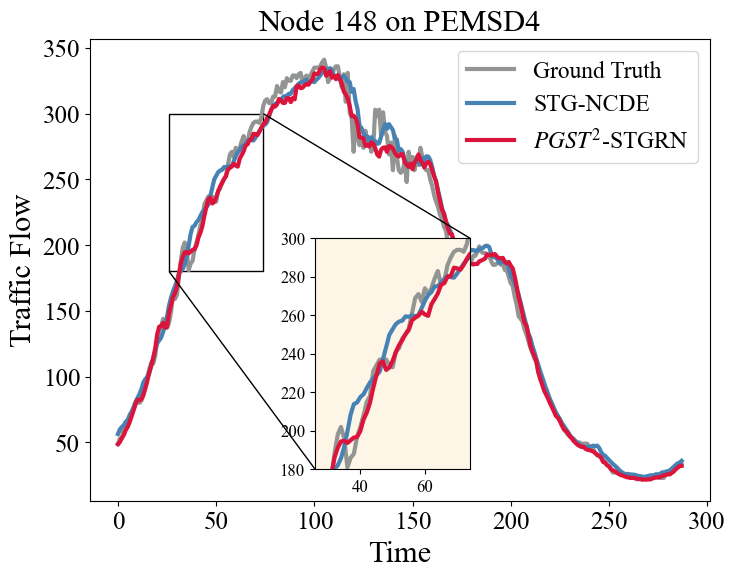}%
\label{fig_visual1}}
\hfil
\subfloat[]{\includegraphics[width=2.2in]{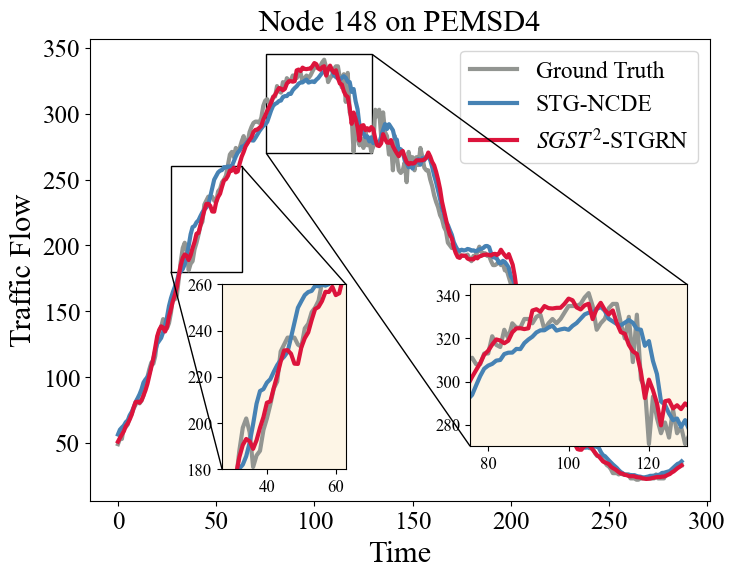}%
\label{fig_visual2}}
\hfil
\subfloat[]{\includegraphics[width=2.2in]{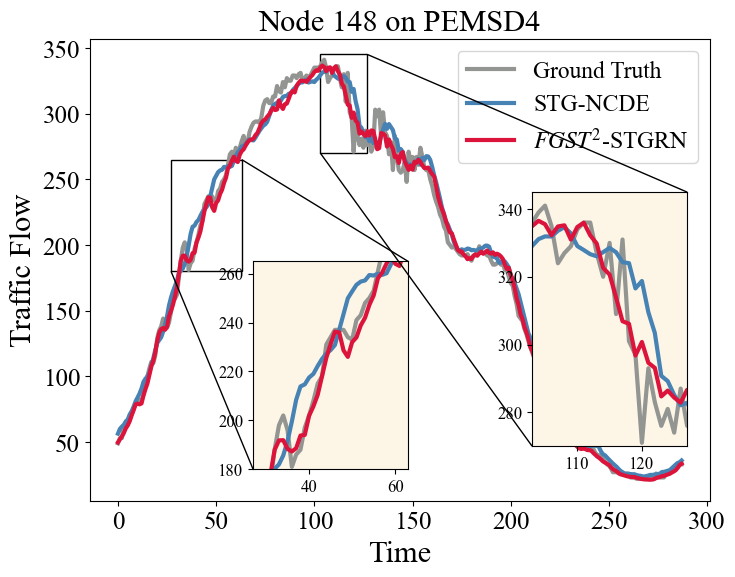}%
\label{fig_visual3}}

\subfloat[]{\includegraphics[width=2.2in]{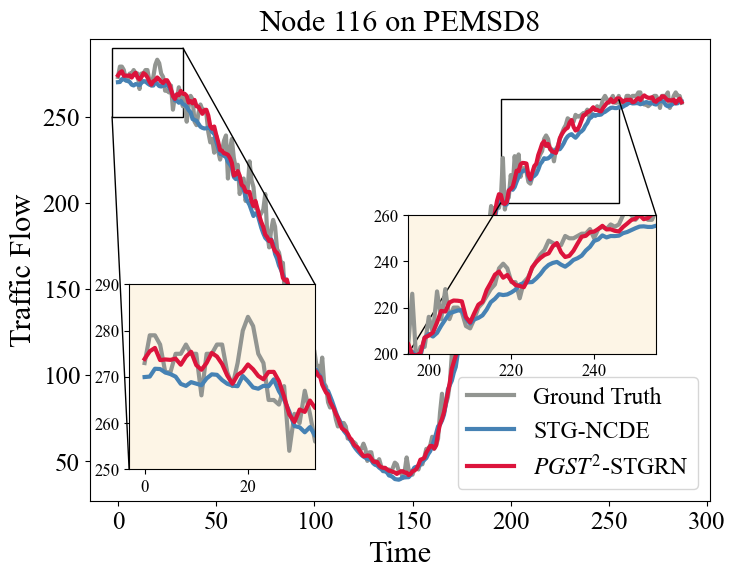}%
\label{fig_visual4}}
\hfil
\subfloat[]{\includegraphics[width=2.2in]{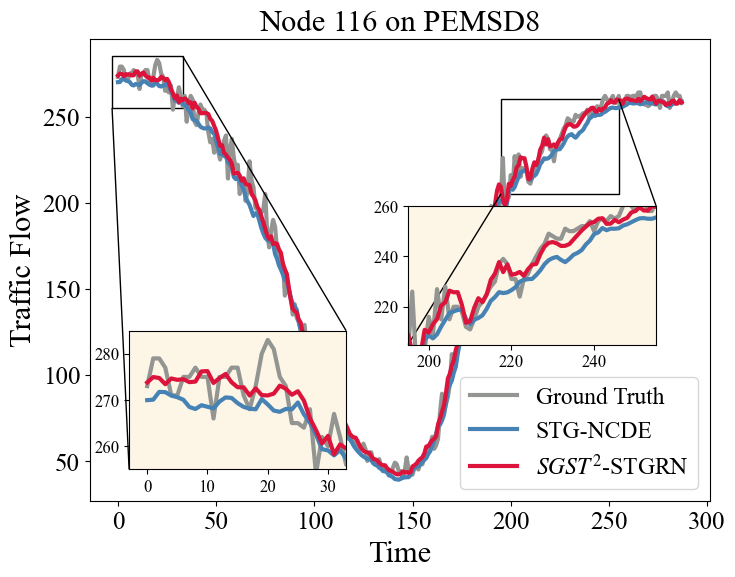}%
\label{fig_visual5}}
\hfil
\subfloat[]{\includegraphics[width=2.2in]{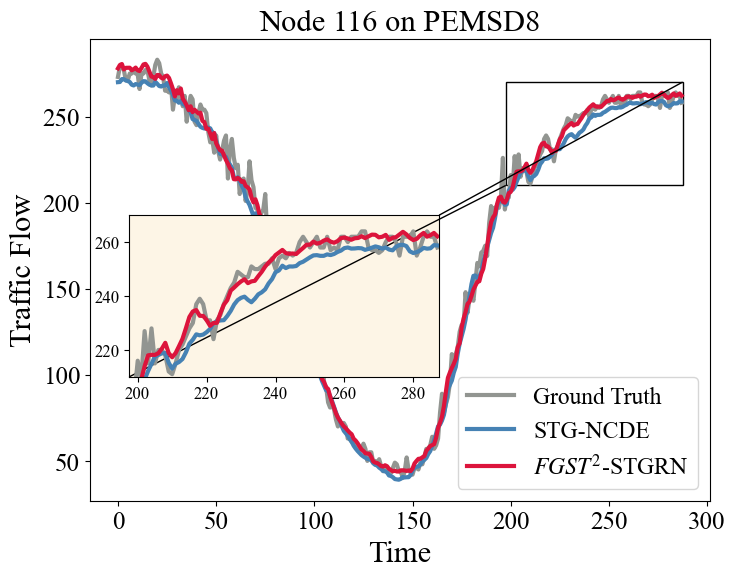}%
\label{fig_visual6}}
\caption{Traffic flow forecasting visualization}
\label{fig_visual}
\end{figure*}

\begin{table*}[t]
	\renewcommand{\arraystretch}{1.2}
	\centering
 \caption{Ablation experiments on \emph{PEMSD4} and \emph{PEMSD8}.}
	\label{ablation}
	\resizebox{\linewidth}{!}{
	\begin{tabular}{c|c c c |c c c c c|c c c|c c c}
            \toprule[1.2pt]
            \multirow{2}{*}{id} & \multicolumn{3}{c|}{\emph{Graph Convolution}} & \multicolumn{5}{c|}{\emph{Global Awareness Layer}} & \multicolumn{3}{c|}{\emph{PEMSD4}} & \multicolumn{3}{c}{\emph{PEMSD8}}\\
		\cline{2-9}
	      {}&{Static Graph}&{Adaptive Graph}&{Sequence-aware Graph}&{TA}&{SA}&{$\mathrm{PGST^2}$}&{$\mathrm{SGST^2}$}&{$\mathrm{FGST^2}$}& \emph{MAE} & \emph{RMSE} & \emph{MAPE} & \emph{MAE} & \emph{RMSE} & \emph{MAPE} \\
		\midrule
		{1}&{\checkmark} & {} & {}& {}& {}& {\checkmark}& {}& {}& 19.70 & 31.98 & 12.88\% & 15.80 & 25.23 & 10.07\%  \\
		{2}&{\checkmark} & {}& {}& {}& {}& {}& {\checkmark}& {}& 19.57 & 31.57 & 12.83\% & 15.62 & 25.04 & 10.10\%  \\
		{3}&{\checkmark} & {}& {}& {}& {}& {}& {}& {\checkmark}& 19.62 & 31.95 & 12.80\% & 15.89 & 25.22 & 10.10\%  \\
		{4}&{}& {\checkmark}& {}& {}& {}& {\checkmark}& {}& {} & 19.25 & 31.75 & 12.79\% & 15.35 & 24.66 & 9.81\%  \\
        {5}&{}& {\checkmark}& {}& {}& {}& {}& {\checkmark}& {} & 19.20 & 31.68 & 12.71\% & 15.21 & 24.84 & 9.77\%  \\
        {6}&{}& {\checkmark}& {}& {}& {}& {}& {}& {\checkmark} & 19.31 & 31.85 & 12.69\% & 15.33 & 24.60 & 10.00\%  \\
        {7}&{}& {}& {\checkmark}& {}& {}& {}& {}& {} & 19.03 & 30.88 & 12.56\% & 16.38 & 27.12 & 10.11\%  \\
        {8}&{}& {}& {\checkmark}& {\checkmark}& {}& {}& {}& {} & 18.95 & 32.08 & 12.39\% & 15.02 & 24.29 & 9.70\%  \\
        {9}&{}& {}& {\checkmark}& {}& {\checkmark}& {}& {}& {} & 20.78 & 34.45 & 13.74\% & 16.92 & 27.54 & 11.06\%  \\
        {10}&{}& {}& {\checkmark}& {}& {}& {\checkmark}& {}& {} & 18.69 & 30.72 & 12.11\% & 14.78 & 24.08 & 9.51\%  \\
        {11}&{}& {}& {\checkmark}& {}& {}& {}& {\checkmark}& {} & 18.71 & 30.64 & 112.33\% & 14.67 & 23.92 & 9.59\%  \\
        {12}&{}& {}& {\checkmark}& {}& {}& {}& {}& {\checkmark} & 18.62 & 30.72 & 12.09\% & 14.70 & 23.97 & 9.64\%  \\
		\bottomrule[1.2pt]
	\end{tabular}
	}
        % \vspace{-0.1in}
	% \caption{Ablation experiments on \emph{PEMSD4} and \emph{PEMSD8}.}
	% \label{ablation}
 % \vspace{-0.1in}
\end{table*}

The traditional statistical methods HA, ARIMA, VAR and SVR have difficulty in handling non-linear data and have poor predictive performance. CNN-based models in existing traffic prediction work have worse or comparable performance compared to RNN-based methods. AGCRN, Z-GCNET and RGSL belong to STGRNs. Although they achieve fairly good prediction performance, our model is much better than them, and the introduction of our $\mathrm{GST^2}$ can enhance their ability to capture global information. Although ST-WA has good predictive performance, its sole reliance on the attention module for spatial-temporal modeling and disregard for the advantages of graph convolution in spatial modeling may somewhat constrain its predictive capabilities. Both STGODE and STG-NCDE use differential equation-based modeling, but their performance is inferior to our proposed models.

Crucially, Table \ref{table_cost} presents the training time, inference time and memory cost of both our models and several recent, high-performing baselines. The evaluation is conducted using the PEMSD4 dataset. The results demonstrate that our models not only achieve superior predictive performance compared to all baseline methods but also maintain highly competitive training and inference times. The expense in the memory size is reasonable considering the continually improving hardware.

We visualise the prediction results and ground truth for $\mathrm{PGST^2}$-STGRN, $\mathrm{SGST^2}$-STGRN, $\mathrm{FGST^2}$-STGRN and STG-NCDE for PEMSD4 and PEMSD8 at 15 min ahead, as shown in Figure \ref{fig_visual}. Node $148$ (or node $116$) is the sensor on the PEMSD4 (or PEMSD8) dataset. While the prediction curves for STG-NCDE are similar to ours at most time points, the highlighted portion of the box shows the stronger predictive performance of our method in challenging situations (e.g., peak periods and high flow fluctuations).

\subsection{Ablation and Parameter Study}
In order to further evaluate the effectiveness of each module and the effect of hyperparameter adjustment, we conducted a series of experiments on the core modules and important parameters of $\mathrm{PGST^2}$-STGRN on the PEMSD4 and PEMSD8.

% \smallskip
% {\noindent \bf Ablation Study.}
\subsubsection{Ablation Study}
We conducted a comprehensive ablation study of three $\mathrm{GST^2}$-STGRNs, mainly two main modules graph convolution and global awareness layer. Among them, the graph convolution can choose from static, adaptive or sequence-aware graphs, and the global awareness layer can choose from temporal attention (TA), Spatial attention (SA), $\mathrm{PGST^2}$, $\mathrm{SGST^2}$ or $\mathrm{FGST^2}$.

Detailed results are shown in the Table \ref{ablation}. Firstly, the adaptive adjacency matrix does outperform the static graph, and the introduction of temporal embedding leads to better spatial modelling. Furthermore, integrating the global aware layer can effectively improve the prediction performance of STGRNs, and temporal attention seems to be more important than spatial attention.

\begin{table}[t]
	\renewcommand{\arraystretch}{1.2}
	\centering
 \caption{Effects of varied convolution depths $K$ on PEMSD8}
	\label{K}
	\resizebox{\columnwidth}{!}{
	\begin{tabular}{c|c|c c c c c c}
		\toprule[1.2pt]
		Model& {$K$}& MAE & RMSE & MAPE & Training & Inference & Memory \\
		\midrule
		\multirow{3}{*}{$\mathrm{\bf{PGST^2}}${\bf{-STGRN}}} & \bf{1} & \bf{14.78} & \bf{24.08} & 9.51\% & \bf{37.60} & \bf{4.80} & \bf{5378} \\
		{} & 2 & 14.87 & 24.28 & 9.52\% & 43.19 & 5.39 & 5844  \\
		{} & 3 & 14.97 & 24.51 & \bf{9.50}\% & 48.25 & 6.04 & 6370 \\
  \midrule
		\multirow{3}{*}{$\mathrm{\bf{SGST^2}}${\bf{-STGRN}}} & \bf{1} & \bf{14.67} & \bf{23.92} & \bf{9.59}\% & \bf{36.70} & \bf{4.58} & \bf{5382} \\
		{} & 2 & 14.70 & 24.14 & 9.58\% & 45.90 & 5.94 & 5844  \\
		{} & 3 & 15.17 & 24.12 & 9.83 & 47.57 & 5.84 & 6400 \\
  \midrule	
		\multirow{3}{*}{$\mathrm{\bf{FGST^2}}${\bf{-STGRN}}} & \bf{1} & \bf{14.70} & \bf{23.97} & \bf{9.64}\% & \bf{37.36} & \bf{4.70} & \bf{5194} \\
		{} & 2 & 15.15 & 24.41 & 9.65\% & 43.27 & 5.42 & 5660  \\
		{} & 3 & 14.92 & 24.30 & 9.71\% & 47.55 & 5.93 & 6184\\
		\bottomrule[1.2pt]
	\end{tabular}
	}
 % \vspace{-0.1in}
	% \caption{Effects of varied convolution depths $K$ on PEMSD8}
	% \label{K}
 % \vspace{-0.2in}
\end{table}

% \smallskip
% {\noindent \bf Hyperparameter Study.}
\subsubsection{Hyperparameter Study}
\begin{figure}[!t]
\centering
\subfloat[]{\includegraphics[width=2.4in]{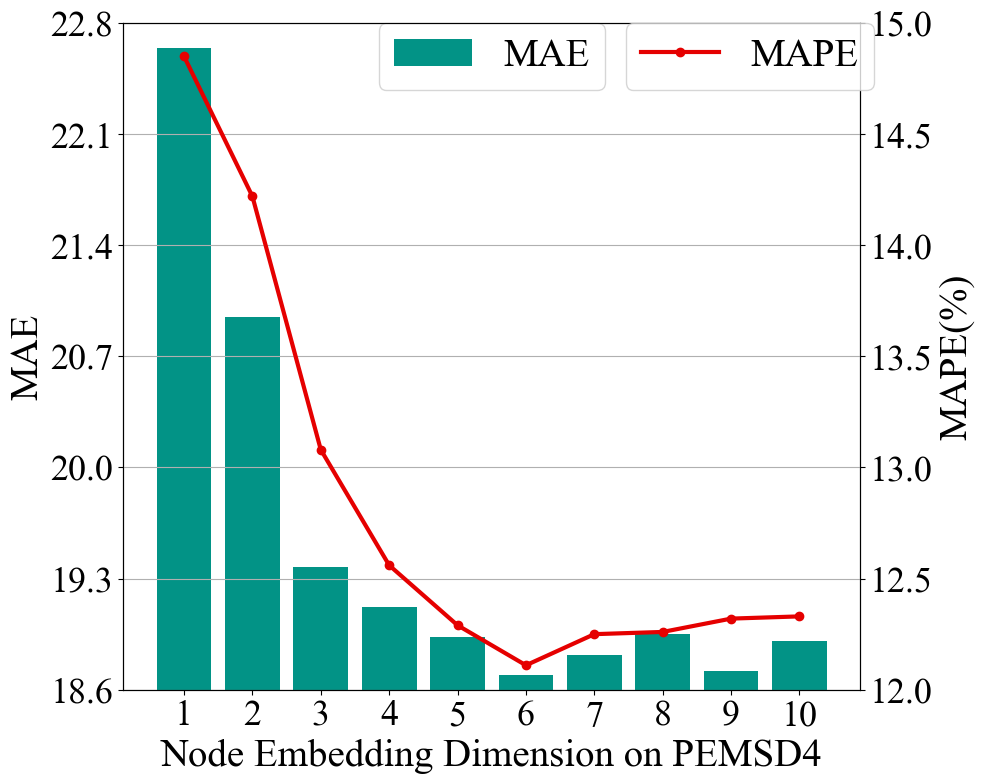}%
\label{e1}}

\subfloat[]{\includegraphics[width=2.4in]{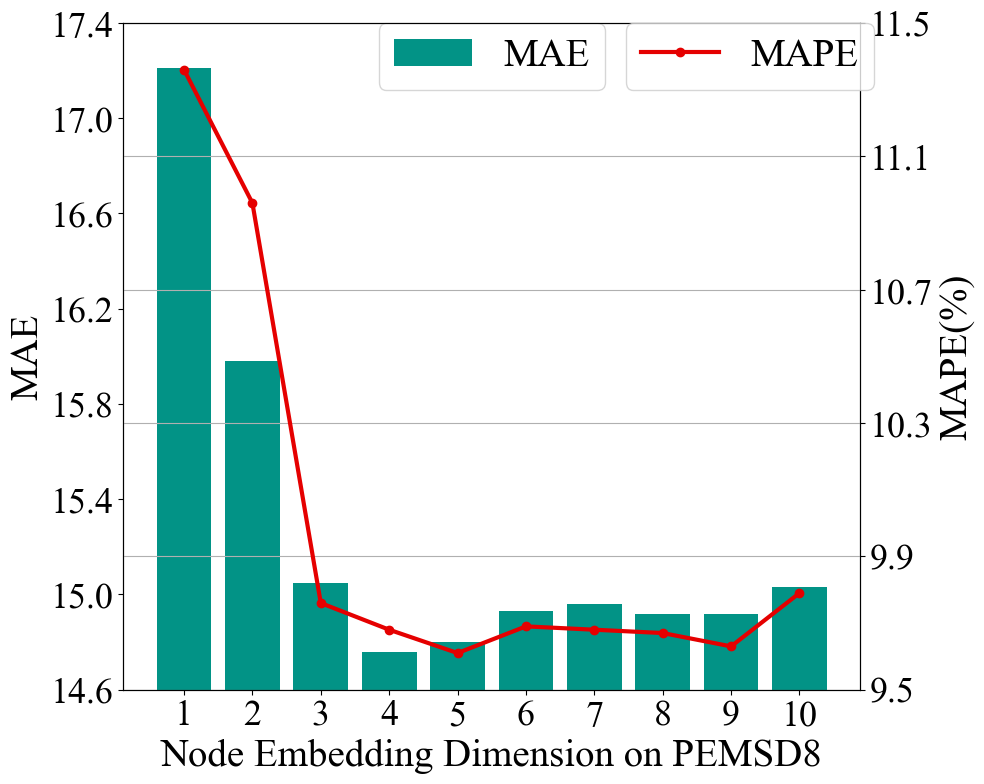}%
\label{e2}}
\caption{Effects of varied node embedding dimension $E_{n}$}
\label{dim}
\end{figure}

Table \ref{K} shows the prediction performance and training cost under different convolution depths $K$. It can be seen that increasing the convolution depth does not improve the prediction performance, but rather increases the training time and memory cost. Therefore, for the experiments we use the 1-order Chebyshev graph convolution network. Meanwhile, Figure \ref{dim} shows effects of varied node embedding dimension on the prediction performance of $\mathrm{PGST^2}$-STGRN. We observe that the node embedding dimension is either too small (underfitting) or too large (overfitting) to impact the model prediction performance, so setting a suitable dimension is critical.

\section{Conclusion}
\noindent This paper presents a pioneering framework called GA-STGRN, that enhances any spatial-temporal graph recurrent neural network with strong global perception. We integrate a sequence-aware graph structure learning module that considers evolving spatial features at different time steps, into parallel, serial, and fused global awareness layers, formulating our 3 concrete models. Empirical findings confirm the superiority of the GA-STGRN framework over the conventional STGRN, and the 3 models' excellent predictive power. We envision that GA-STGRN can influence the design and development of spatial-temporal data mining techniques.

\section*{Acknowledgments}
Chunjiang Zhu is supported by UNCG Start-up Funds and Faculty First Award. Detian Zhang is partially supported by the Collaborative Innovation Center of Novel Software Technology and Industrialization, the Priority Academic Program Development of Jiangsu Higher Education Institutions. 
%Jin Wang is supported in part by  National Natural Science Foundation of China (62072321), Six Talent Peak Project of Jiangsu Province (XYDXX-084), Natural Science Foundation of the Higher Education Institutions of Jiangsu Province (22KJA520007)

% \bf{If you will not include a photo:}\vspace{-33pt}
% \begin{IEEEbiographynophoto}{John Doe}
% Use $\backslash${\tt{begin\{IEEEbiographynophoto\}}} and the author name as the argument followed by the biography text.
% \end{IEEEbiographynophoto}
\bibliographystyle{IEEEtran}
\bibliography{tits}

% Generated by IEEEtran.bst, version: 1.14 (2015/08/26)
\begin{thebibliography}{10}
\providecommand{\url}[1]{#1}
\csname url@samestyle\endcsname
\providecommand{\newblock}{\relax}
\providecommand{\bibinfo}[2]{#2}
\providecommand{\BIBentrySTDinterwordspacing}{\spaceskip=0pt\relax}
\providecommand{\BIBentryALTinterwordstretchfactor}{4}
\providecommand{\BIBentryALTinterwordspacing}{\spaceskip=\fontdimen2\font plus
\BIBentryALTinterwordstretchfactor\fontdimen3\font minus \fontdimen4\font\relax}
\providecommand{\BIBforeignlanguage}[2]{{%
\expandafter\ifx\csname l@#1\endcsname\relax
\typeout{** WARNING: IEEEtran.bst: No hyphenation pattern has been}%
\typeout{** loaded for the language `#1'. Using the pattern for}%
\typeout{** the default language instead.}%
\else
\language=\csname l@#1\endcsname
\fi
#2}}
\providecommand{\BIBdecl}{\relax}
\BIBdecl

\bibitem{wu2020comprehensive}
Z.~Wu, S.~Pan, F.~Chen, G.~Long, C.~Zhang, and S.~Y. Philip, ``A comprehensive survey on graph neural networks,'' \emph{IEEE transactions on neural networks and learning systems}, vol.~32, no.~1, pp. 4--24, 2020.

\bibitem{jiang2021dl}
R.~Jiang, D.~Yin, Z.~Wang, Y.~Wang, J.~Deng, H.~Liu, Z.~Cai, J.~Deng, X.~Song, and R.~Shibasaki, ``Dl-traff: Survey and benchmark of deep learning models for urban traffic prediction,'' in \emph{Proceedings of the 30th ACM international conference on information \& knowledge management}, 2021, pp. 4515--4525.

\bibitem{sutskever2014sequence}
I.~Sutskever, O.~Vinyals, and Q.~V. Le, ``Sequence to sequence learning with neural networks,'' \emph{Advances in neural information processing systems}, vol.~27, 2014.

\bibitem{li2018dcrnn_traffic}
Y.~Li, R.~Yu, C.~Shahabi, and Y.~Liu, ``Diffusion convolutional recurrent neural network: Data-driven traffic forecasting,'' in \emph{International Conference on Learning Representations (ICLR '18)}, 2018.

\bibitem{bai2020adaptive}
L.~Bai, L.~Yao, C.~Li, X.~Wang, and C.~Wang, ``Adaptive graph convolutional recurrent network for traffic forecasting,'' \emph{Advances in Neural Information Processing Systems}, vol.~33, pp. 17\,804--17\,815, 2020.

\bibitem{chen2021z}
Y.~Chen, I.~Segovia, and Y.~R. Gel, ``Z-gcnets: time zigzags at graph convolutional networks for time series forecasting,'' in \emph{International Conference on Machine Learning}.\hskip 1em plus 0.5em minus 0.4em\relax PMLR, 2021, pp. 1684--1694.

\bibitem{yu2022regularized}
H.~Yu, T.~Li, W.~Yu, J.~Li, Y.~Huang, L.~Wang, and A.~Liu, ``Regularized graph structure learning with semantic knowledge for multi-variates time-series forecasting,'' in \emph{Proceedings of the Thirty-First International Joint Conference on Artificial Intelligence, {IJCAI-22}}, 2022, pp. 2362--2368.

\bibitem{liu2022msdr}
D.~Liu, J.~Wang, S.~Shang, and P.~Han, ``Msdr: Multi-step dependency relation networks for spatial temporal forecasting,'' in \emph{Proceedings of the 28th ACM SIGKDD Conference on Knowledge Discovery and Data Mining}, 2022, pp. 1042--1050.

\bibitem{yu2018spatio}
B.~Yu, H.~Yin, and Z.~Zhu, ``Spatio-temporal graph convolutional networks: a deep learning framework for traffic forecasting,'' in \emph{Proceedings of the 27th International Joint Conference on Artificial Intelligence}, 2018, pp. 3634--3640.

\bibitem{wu2019graph}
Z.~Wu, S.~Pan, G.~Long, J.~Jiang, and C.~Zhang, ``Graph wavenet for deep spatial-temporal graph modeling,'' in \emph{Proceedings of the 28th International Joint Conference on Artificial Intelligence}, 2019, pp. 1907--1913.

\bibitem{huang2020lsgcn}
R.~Huang, C.~Huang, Y.~Liu, G.~Dai, and W.~Kong, ``Lsgcn: Long short-term traffic prediction with graph convolutional networks.'' in \emph{IJCAI}, 2020, pp. 2355--2361.

\bibitem{song2020spatial}
C.~Song, Y.~Lin, S.~Guo, and H.~Wan, ``Spatial-temporal synchronous graph convolutional networks: A new framework for spatial-temporal network data forecasting,'' in \emph{Proceedings of the AAAI Conference on Artificial Intelligence}, vol.~34, no.~01, 2020, pp. 914--921.

\bibitem{li2021spatial}
M.~Li and Z.~Zhu, ``Spatial-temporal fusion graph neural networks for traffic flow forecasting,'' in \emph{Proceedings of the AAAI conference on artificial intelligence}, vol.~35, no.~5, 2021, pp. 4189--4196.

\bibitem{guo2019attention}
S.~Guo, Y.~Lin, N.~Feng, C.~Song, and H.~Wan, ``Attention based spatial-temporal graph convolutional networks for traffic flow forecasting,'' in \emph{Proceedings of the AAAI conference on artificial intelligence}, vol.~33, no.~01, 2019, pp. 922--929.

\bibitem{wang2020traffic}
X.~Wang, Y.~Ma, Y.~Wang, W.~Jin, X.~Wang, J.~Tang, C.~Jia, and J.~Yu, ``Traffic flow prediction via spatial temporal graph neural network,'' in \emph{Proceedings of The Web Conference 2020}, 2020, pp. 1082--1092.

\bibitem{zheng2020gman}
C.~Zheng, X.~Fan, C.~Wang, and J.~Qi, ``Gman: A graph multi-attention network for traffic prediction,'' in \emph{Proceedings of the AAAI conference on artificial intelligence}, vol.~34, no.~01, 2020, pp. 1234--1241.

\bibitem{lan2022dstagnn}
S.~Lan, Y.~Ma, W.~Huang, W.~Wang, H.~Yang, and P.~Li, ``Dstagnn: Dynamic spatial-temporal aware graph neural network for traffic flow forecasting,'' in \emph{International Conference on Machine Learning}.\hskip 1em plus 0.5em minus 0.4em\relax PMLR, 2022, pp. 11\,906--11\,917.

\bibitem{cirstea2022towards}
R.-G. Cirstea, B.~Yang, C.~Guo, T.~Kieu, and S.~Pan, ``Towards spatio-temporal aware traffic time series forecasting,'' in \emph{2022 IEEE 38th International Conference on Data Engineering (ICDE)}.\hskip 1em plus 0.5em minus 0.4em\relax IEEE, 2022, pp. 2900--2913.

\bibitem{chen2021tamp}
Y.~Chen, I.~Segovia-Dominguez, B.~Coskunuzer, and Y.~Gel, ``Tamp-s2gcnets: coupling time-aware multipersistence knowledge representation with spatio-supra graph convolutional networks for time-series forecasting,'' in \emph{International Conference on Learning Representations}, 2021.

\bibitem{chen2022time}
Y.~Chen, Y.~Gel, and H.~V. Poor, ``Time-conditioned dances with simplicial complexes: Zigzag filtration curve based supra-hodge convolution networks for time-series forecasting,'' \emph{Advances in Neural Information Processing Systems}, vol.~35, pp. 8940--8953, 2022.

\bibitem{zivot2006vector}
E.~Zivot and J.~Wang, ``Vector autoregressive models for multivariate time series,'' \emph{Modeling financial time series with S-PLUS{\textregistered}}, pp. 385--429, 2006.

\bibitem{lee1999application}
S.~Lee and D.~B. Fambro, ``Application of subset autoregressive integrated moving average model for short-term freeway traffic volume forecasting,'' \emph{Transportation research record}, vol. 1678, no.~1, pp. 179--188, 1999.

\bibitem{williams2003modeling}
B.~M. Williams and L.~A. Hoel, ``Modeling and forecasting vehicular traffic flow as a seasonal arima process: Theoretical basis and empirical results,'' \emph{Journal of transportation engineering}, vol. 129, no.~6, pp. 664--672, 2003.

\bibitem{drucker1996support}
H.~Drucker, C.~J. Burges, L.~Kaufman, A.~Smola, and V.~Vapnik, ``Support vector regression machines,'' \emph{Advances in neural information processing systems}, vol.~9, 1996.

\bibitem{wu2004travel}
C.-H. Wu, J.-M. Ho, and D.-T. Lee, ``Travel-time prediction with support vector regression,'' \emph{IEEE transactions on intelligent transportation systems}, vol.~5, no.~4, pp. 276--281, 2004.

\bibitem{defferrard2016convolutional}
M.~Defferrard, X.~Bresson, and P.~Vandergheynst, ``Convolutional neural networks on graphs with fast localized spectral filtering,'' \emph{Advances in neural information processing systems}, vol.~29, 2016.

\bibitem{kipf2016semi}
T.~N. Kipf and M.~Welling, ``Semi-supervised classification with graph convolutional networks,'' \emph{arXiv preprint arXiv:1609.02907}, 2016.

\bibitem{fang2021spatial}
Z.~Fang, Q.~Long, G.~Song, and K.~Xie, ``Spatial-temporal graph ode networks for traffic flow forecasting,'' in \emph{Proceedings of the 27th ACM SIGKDD Conference on Knowledge Discovery \& Data Mining}, 2021, pp. 364--373.

\bibitem{choi2022graph}
J.~Choi, H.~Choi, J.~Hwang, and N.~Park, ``Graph neural controlled differential equations for traffic forecasting,'' in \emph{Proceedings of the AAAI Conference on Artificial Intelligence}, vol.~36, no.~6, 2022, pp. 6367--6374.

\bibitem{hechtlinger2017generalization}
Y.~Hechtlinger, P.~Chakravarti, and J.~Qin, ``A generalization of convolutional neural networks to graph-structured data,'' \emph{arXiv preprint arXiv:1704.08165}, 2017.

\bibitem{hamilton2017inductive}
W.~Hamilton, Z.~Ying, and J.~Leskovec, ``Inductive representation learning on large graphs,'' \emph{Advances in neural information processing systems}, vol.~30, 2017.

\bibitem{velivckovic2017graph}
P.~Veli{\v{c}}kovi{\'c}, G.~Cucurull, A.~Casanova, A.~Romero, P.~Lio, and Y.~Bengio, ``Graph attention networks,'' \emph{arXiv preprint arXiv:1710.10903}, 2017.

\bibitem{bruna2013spectral}
J.~Bruna, W.~Zaremba, A.~Szlam, and Y.~LeCun, ``Spectral networks and locally connected networks on graphs,'' \emph{arXiv preprint arXiv:1312.6203}, 2013.

\bibitem{radford2019language}
A.~Radford, J.~Wu, R.~Child, D.~Luan, D.~Amodei, I.~Sutskever \emph{et~al.}, ``Language models are unsupervised multitask learners,'' \emph{OpenAI blog}, vol.~1, no.~8, p.~9, 2019.

\bibitem{bai2021segatron}
H.~Bai, P.~Shi, J.~Lin, Y.~Xie, L.~Tan, K.~Xiong, W.~Gao, and M.~Li, ``Segatron: Segment-aware transformer for language modeling and understanding,'' in \emph{Proceedings of the AAAI Conference on Artificial Intelligence}, vol.~35, no.~14, 2021, pp. 12\,526--12\,534.

\bibitem{lin2021end}
K.~Lin, L.~Wang, and Z.~Liu, ``End-to-end human pose and mesh reconstruction with transformers,'' in \emph{Proceedings of the IEEE/CVF Conference on Computer Vision and Pattern Recognition}, 2021, pp. 1954--1963.

\bibitem{vaswani2017attention}
A.~Vaswani, N.~Shazeer, N.~Parmar, J.~Uszkoreit, L.~Jones, A.~N. Gomez, {\L}.~Kaiser, and I.~Polosukhin, ``Attention is all you need,'' \emph{Advances in neural information processing systems}, vol.~30, 2017.

\bibitem{chen2001freeway}
C.~Chen, K.~Petty, A.~Skabardonis, P.~Varaiya, and Z.~Jia, ``Freeway performance measurement system: mining loop detector data,'' \emph{Transportation Research Record}, vol. 1748, no.~1, pp. 96--102, 2001.

\end{thebibliography}

\begin{IEEEbiography}[{\includegraphics[width=1in,height=1.25in,clip,keepaspectratio]{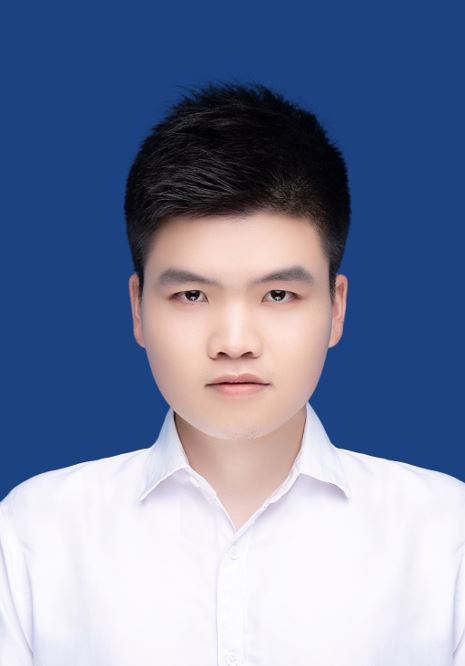}}]{Haiyang Liu}
is currently pursuing the master's degree with the Institute of Artificial Intelligence, Department of Computer Science and Technology, Soochow University, Suzhou, China. His research interests include spatial-temporal databases and intelligent transportation systems.
\end{IEEEbiography}
\begin{IEEEbiography}[{\includegraphics[width=1in,height=1.25in,clip,keepaspectratio]{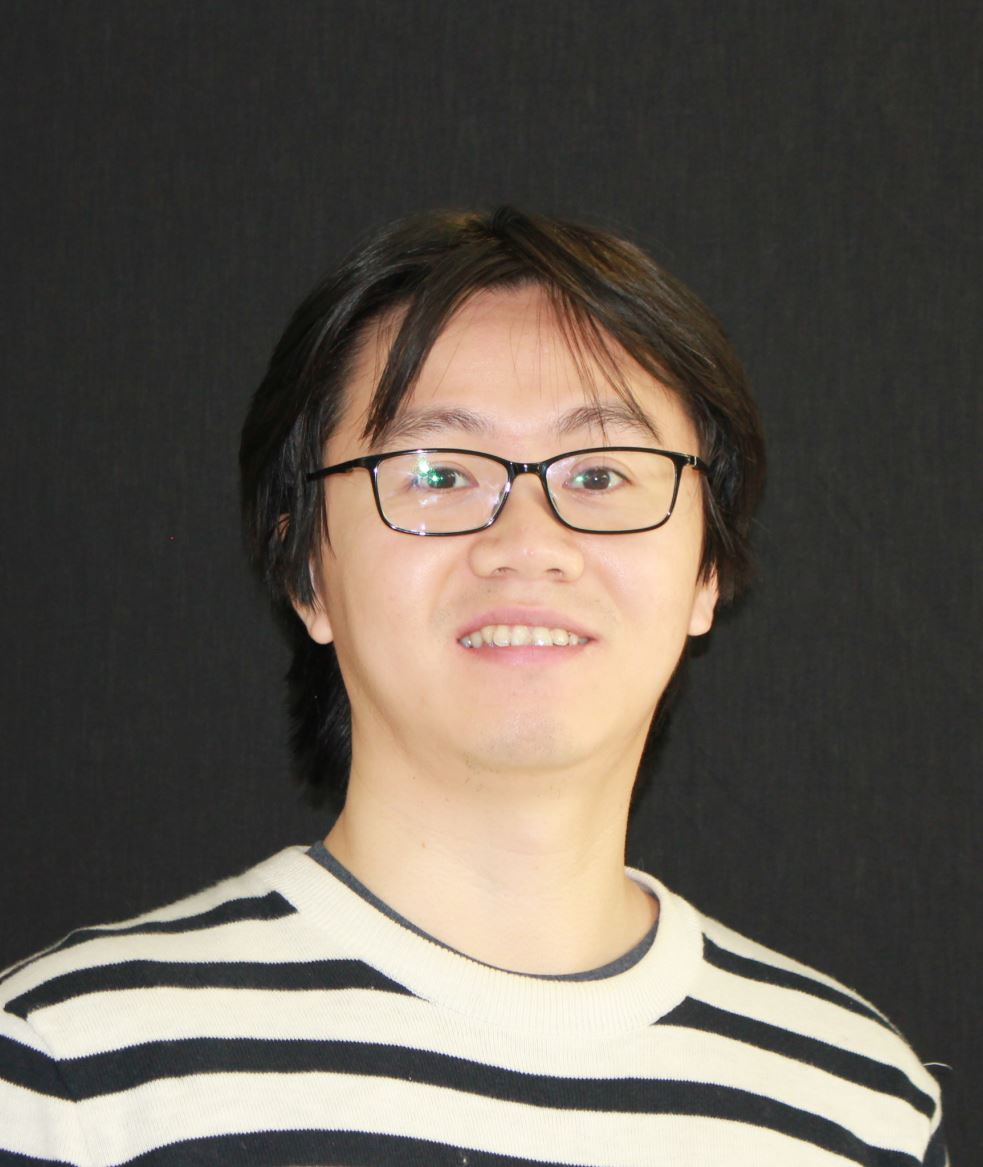}}]{Chunjiang Zhu}
 is an assistant professor in the Department of Computer Science at the University of North Carolina at Greensboro. He received his Ph.D. and Master in Computer Science from City University of Hong Kong and Chinese Academy of Sciences, respectively. His research interests include Machine Learning and Theory, Graph Algorithms, Chemoinformatics, and Cyber-Physical Systems.
\end{IEEEbiography}
\begin{IEEEbiography}[{\includegraphics[width=1in,height=1.25in,clip,keepaspectratio]{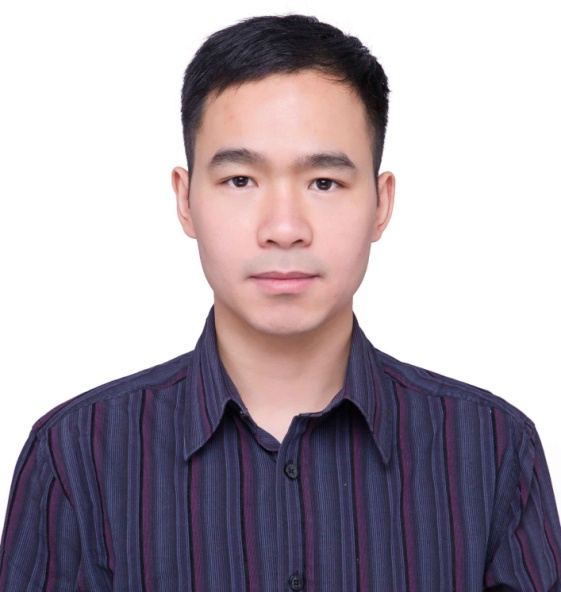}}]{Detian Zhang}
received the Ph.D. degree in computer science from City University of Hong Kong and University of Science and Technology of China in 2014. He is currently an associate professor with the Institute of Artificial Intelligence, School of Computer Science and Technology, Soochow University, Suzhou, China. His research interests include spatial-temporal databases and intelligent transportation systems.
\end{IEEEbiography}
% \begin{IEEEbiography}[{\includegraphics[width=1in,height=1.25in,clip, keepaspectratio]{figures/authorJin_WANG.eps}}] {Jin Wang}(M'12) received the B.S. degree from Ocean University of China in 2006, and the Ph.D. degree in computer science jointly awarded by City University of Hong Kong and University of Science and Technology of China in 2011. He is currently a professor at the Department of Computer Science and Technology, Soochow University, Suzhou, China. His research interests include edge computing and network security.
% \end{IEEEbiography}
\vfill

\end{document}